\documentclass[acmsmall]{acmart}

\AtBeginDocument{%
  \providecommand\BibTeX{{%
    \normalfont B\kern-0.5em{\scshape i\kern-0.25em b}\kern-0.8em\TeX}}}

\acmJournal{CSUR}

\usepackage{amsmath,amsfonts}
\usepackage{mathtools}
\usepackage{array}
\usepackage{textcomp}
\usepackage{stfloats}
\usepackage{url}
\usepackage{verbatim}
\usepackage{graphicx}

\usepackage{amssymb}

\usepackage{amscd}
\usepackage{float}
\usepackage{placeins}
\usepackage{subfig}
\usepackage{nccmath}
\usepackage{siunitx}
\usepackage{booktabs}
\usepackage{tikz}
\usetikzlibrary{arrows.meta,positioning,shapes.geometric}
\usepackage{url}

\begin{document}

\title{A Tutorial on Gaussian Process Learning-based Model Predictive Control}

\author{Jie Wang}
\email{j.wang@umanitoba.ca}
\orcid{0000-0001-9970-1808}
\authornote{Corresponding author.}
\affiliation{%
  \institution{University of Manitoba}
  \streetaddress{75 Chancellors Circle}
  \city{Winnipeg}
  \state{MB}
  \country{Canada}
  \postcode{R3T 5V6}
}

\author{Youmin Zhang}
\email{youmin.zhang@concordia.ca
}
\affiliation{%
  \institution{Concordia University}
  \streetaddress{1455 Blvd. De Maisonneuve Ouest}
  \city{Montreal}
  \state{QC}
  \country{Canada}
  \postcode{H3G 1M8}
}

\renewcommand{\shortauthors}{J. Wang et al.}

\begin{abstract}
This tutorial provides an accessible and implementation-oriented introduction to Gaussian process learning-based model predictive control (GP-MPC), which combines probabilistic residual modeling with receding-horizon control. Its central tutorial contribution is a detailed, step-by-step derivation of multi-step mean and covariance propagation for GP-augmented prediction models. The derivation shows how commonly used propagation formulas follow from the laws of total expectation and total covariance, and clarifies the roles of the GP posterior mean, GP posterior covariance, uncertain inputs, and query--output cross-covariances. Building on this foundation, the paper distinguishes mean-only unconstrained GP predictive control from uncertainty-aware constrained GP-MPC, clarifies regulation and output-tracking formulations, and provides concise implementation and computational-complexity guidance, including practical software-tool references and discussion of the scope of closed-loop guarantees. Mobile-robot path-following examples illustrate mean-only unconstrained implementations, whereas mixed-vehicle platooning illustrates uncertainty-aware constrained GP-MPC with uncertainty-dependent safety-distance tightening. The tutorial is intended to help researchers and practitioners understand, implement, and critically evaluate GP-MPC designs for robotic and autonomous systems operating under modeling uncertainty.
\end{abstract}

\ccsdesc[500]{Computer systems organization~Robotics}
\ccsdesc[500]{Computing methodologies~Gaussian processes}

\keywords{Gaussian process, model predictive control, learning-based control, mobile robotics, dynamic modeling}


\maketitle

\section{Introduction}
\label{sec:intro}
Robotic and autonomous systems increasingly operate in environments where physics-based nominal models are useful but incomplete: mobile robots encounter changing terrain conditions, autonomous vehicles interact with human drivers, and model mismatch can vary over the operating domain \cite{brunke2022safe}. Model predictive control (MPC) is well suited to these settings because it repeatedly optimizes predicted behavior over a finite horizon while providing a direct mechanism for including input, state, and output constraints \cite{hewing2020learning,deshpande2021fault,chevet2020decentralized}. Gaussian process (GP) regression complements this model-based structure by learning residual dynamics from data and returning both a posterior mean and a posterior covariance that can represent model uncertainty \cite{liu2018gaussian,wang2023intuitive}.

Historically, MPC developed as a receding-horizon model-based control methodology in which control actions are determined online from constrained optimization. GPs developed as non-parametric probabilistic models that use covariance functions to represent distributions over unknown functions. GP-MPC combines these ideas: a nominal model retains physical structure and interpretability, whereas a GP residual model improves predictions in regions supported by data and can provide uncertainty information for cautious constraint handling. This combination is attractive when a complete first-principles model is difficult to obtain but data alone are insufficient for safely controlling a system \cite{brunke2022safe}.

GP-MPC should also be distinguished from policy-learning approaches such as reinforcement learning (RL). GP-MPC retains an explicit predictive model, an online receding-horizon optimization problem, and direct constraint representations. RL typically seeks a policy or value function from environmental interactions or data, potentially reducing online optimization after training, but constraint satisfaction and safety commonly require additional design mechanisms \cite{brunke2022safe}. These paradigms are therefore complementary: learning can improve models or policies, while MPC provides an optimization-based control architecture with explicit physical constraints.

Recent reviews and tutorials have provided valuable perspectives on learning-based MPC \cite{hewing2020learning}, GP-based control \cite{liu2018gaussian}, and safe learning in robotics \cite{brunke2022safe}. More recent survey work focused specifically on GP dynamics learning in MPC further identifies scalability, tractable prediction approximations, online model adaptation, and theoretical guarantees as central design issues \cite{scampicchio2025gaussian}. Despite these important contributions, the detailed mathematical steps needed to integrate GP predictions into multi-step MPC predictions are often compressed or omitted in research papers. This tutorial addresses that gap by providing an equation-by-equation derivation of mean and covariance propagation for GP-augmented multi-step prediction, thereby making the underlying GP-MPC formulation more accessible and reproducible.

A central mathematical contribution of this tutorial is the derivation in Sec.~\ref{sec:mean_var_approx}. Instead of only presenting the final GP-MPC prediction equations, this section starts from the GP-augmented state-space model, applies the laws of total expectation and total covariance, and shows how Taylor and mean-equivalent covariance approximations lead to MPC-ready formulas. This presentation makes the assumptions, neglected terms, and computational trade-offs transparent for readers who wish to implement or modify GP-MPC algorithms.

The central implementation distinction used throughout this tutorial is between two controller families. \emph{Mean-only unconstrained GP-MPC} incorporates the GP posterior mean into the prediction model to improve nominal tracking predictions, but it neither propagates GP covariance nor imposes explicit online inequality constraints in the reported implementations. \emph{Uncertainty-aware constrained GP-MPC} propagates predictive uncertainty and uses it in explicit constraints, for example by tightening a probabilistic safety-distance constraint. This distinction is important because GP-enhanced prediction alone does not establish constrained MPC behavior or formal closed-loop guarantees.

The tutorial makes four contributions. First, and most centrally, it provides a tutorial-style, step-by-step derivation of tractable mean and covariance propagation approximations for GP-augmented multi-step prediction. Second, it clearly formulates regulation and output-tracking problems and distinguishes unconstrained mean-only and constrained uncertainty-aware GP-MPC. Third, it provides concise implementation and computational-complexity guidance, references practical software tools, and discusses the scope of stability and feasibility guarantees. Fourth, it maps the formulations to applications: mobile-robot path following demonstrates mean-only unconstrained GP predictive control, whereas mixed-vehicle platooning demonstrates constrained GP-MPC with uncertainty-dependent safe-distance tightening.

The remainder of this paper is organized as follows. Sec.~\ref{sec:gp} summarizes the GP residual-model notation needed for the GP-MPC derivations. Sec.~\ref{sec:mean_var_approx} develops approximations for multi-step mean and uncertainty propagation. Sec.~\ref{sec:gp_mpc} defines the GP-MPC problem formulations, concise implementation guidance, and the scope of closed-loop guarantees. Secs.~\ref{sec:path_following} and \ref{sec:mixed_platoon} present the mobile-robot path-following and mixed-vehicle platooning applications, respectively. Finally, Sec.~\ref{sec:conclusion} summarizes the main lessons and limitations.

\section{Gaussian Process Notation Used in GP-MPC}
\label{sec:gp}
This section summarizes only the Gaussian process (GP) notation needed for the GP-MPC derivations in Sec.~\ref{sec:mean_var_approx}. More elementary introductions to GP regression, including visual explanations of Gaussian distributions, kernels, hyperparameter effects, and predictive uncertainty, are available in the GP tutorial literature \cite{wang2023intuitive}. Here, we focus on the residual-modeling quantities that enter the predictive-control formulation.

Let $z\in\mathbb{R}^{n_z}$ denote a GP query, typically formed by stacking the system state and input, $z=[x^\top,u^\top]^\top$. A GP residual model represents an unknown function $d(z)$ by a posterior distribution conditioned on training data $\mathcal{D}$,
\begin{equation}
    d(z)\mid\mathcal{D}\sim\mathcal{N}\bigl(\mu^d(z),\Sigma^d(z)\bigr).
    \label{eqn:gp_posterior_short}
\end{equation}
The posterior mean $\mu^d(z)$ provides a learned correction to the nominal model, while the posterior covariance $\Sigma^d(z)$ quantifies uncertainty at the query. For deterministic queries, standard GP regression gives closed-form posterior mean and covariance expressions based on the selected kernel and training data. For multi-step GP-MPC, however, later prediction steps involve uncertain future states. Therefore, the query $z$ itself becomes random, and the quantities $\mu^d(z)$ and $\Sigma^d(z)$ become nonlinear functions of an uncertain input. This is the reason that moment propagation approximations are required in Sec.~\ref{sec:mean_var_approx}.

The computational cost of GP prediction is also important in MPC because the GP may be evaluated many times across the prediction horizon and within solver iterations. For $n$ training points, exact GP model preparation typically requires $\mathcal{O}(n^3)$ operations and $\mathcal{O}(n^2)$ storage, while straightforward single-query prediction requires approximately $\mathcal{O}(n)$ operations for the posterior mean and $\mathcal{O}(n^2)$ operations for the posterior covariance \cite{rasmussen2006,liu2020gaussian}. Sparse GP methods reduce this burden by representing the training information through $m\ll n$ inducing variables, leading to typical online mean and covariance prediction costs of approximately $\mathcal{O}(m)$ and $\mathcal{O}(m^2)$, respectively \cite{liu2020gaussian,snelson2005sparse}. These scaling considerations motivate the sparse and dynamic sparse GP-MPC variants discussed later in Sec.~\ref{sec:sparse_gp_mpc}.

\section{Approximating Mean and Uncertainty Propagation}
\label{sec:mean_var_approx}
This section is the mathematical core of the tutorial. A one-step GP posterior evaluated at a deterministic query is Gaussian, but this property does not automatically carry over to multi-step GP-MPC prediction. After the first prediction step, the predicted state is uncertain; when this uncertain state is used as the next GP input, the resulting GP output distribution is generally non-Gaussian \cite{quinonero2003}. Therefore, practical GP-MPC algorithms usually use a Gaussian closure: at each prediction step, only the first two moments are retained and the next state is approximated as Gaussian again. This section explains, step by step, how those mean and covariance recursion formulas are obtained.



Consider a nonlinear dynamical system represented by a discrete-time state-space model \cite{hewing2018cautious}:
\begin{ceqn}
    \begin{align} 
        x_{k+1}
        &=\overbrace{f(x_k,u_k)}^{\text{prior nominal model}}
        +\overbrace{g(x_k,u_k)}^{\text{learning-based model}} \, , \label{eqn:process_model_general} \\ 
        &= f(x_k,u_k)+B_d\left(d(x_k,u_k)+\omega_k\right) \, . \label{eqn:process_model_Bd} 
    \end{align}
\end{ceqn}
Here, $x_k\in\mathbb{R}^{n_x}$ and $u_k\in\mathbb{R}^{n_u}$ denote the state and control input at time step $k$, respectively. The function $f(x_k,u_k)$ is the physics-based nominal model, which captures the part of the dynamics that is known from modeling. The term $g(x_k,u_k)$ represents the residual dynamics, model mismatch, or difficult-to-model effects that are learned from data. In \eqref{eqn:process_model_Bd}, this residual is represented by a GP model $d(\cdot)\in\mathbb{R}^{n_d}$ and a selection matrix $B_d\in\mathbb{R}^{n_x\times n_d}$. The matrix $B_d$ maps the learned residual into the state equation and can be used to let the GP affect only selected state channels. This is useful when only part of the dynamics is uncertain or when reducing the GP output dimension improves computational efficiency. If residual learning is required for all state channels, $B_d$ can be set to the identity matrix. The term $\omega_k\sim\mathcal{N}(0,\Sigma^\omega)$ denotes zero-mean process noise, where $\Sigma^\omega=\operatorname{diag}([\sigma_1^2,\ldots,\sigma_{n_d}^2])$.

For compact notation, define the GP query vector and the residual output as
\begin{equation*}
    z_k=\begin{bmatrix}x_k^\top & u_k^\top\end{bmatrix}^{\top},
    \qquad
    d_k=d(z_k)=d(x_k,u_k).
\end{equation*}
At a deterministic query $z_k$, the GP posterior of the residual model is
\begin{equation}
    d(z_k)\mid\mathcal{D}
    \sim
    \mathcal{N}\!\left(\mu^d(z_k),\Sigma^d(z_k)\right) \, .\label{eqn:gp_Bd}
\end{equation}
The notation $\mu^d(z_k)$ and $\Sigma^d(z_k)$ denotes the conditional GP posterior mean and covariance evaluated at the query $z_k$. A key point for multi-step GP-MPC is that, once $z_k$ is uncertain, $\mu^d(z_k)$ and $\Sigma^d(z_k)$ are no longer fixed matrices; they become nonlinear functions of a random query. This is the reason why moment approximation is required.

To make multi-step prediction tractable, the predicted state and query distributions are approximated as Gaussian at each future prediction step:
\begin{equation*}
    x_k\sim\mathcal{N}(\mu_k^x,\Sigma_k^x),\qquad
    u_k\sim\mathcal{N}(\mu_k^u,\Sigma_k^u),\qquad
    z_k\sim\mathcal{N}(\mu_k^z,\Sigma_k^z).
\end{equation*}
In a standard open-loop MPC optimization, the future input sequence is usually treated as a deterministic decision variable during one optimizer evaluation; in that case, $\Sigma_k^u=0$ and $\Sigma_k^{xu}=0$. The more general notation is retained here because it makes the propagation formulas applicable to stochastic MPC, feedback-policy MPC, and other settings where the future input may also be uncertain.

Under this Gaussian approximation, the state, input, and residual-plus-noise term can be summarized by the following approximate joint Gaussian distribution:
\begin{ceqn}
\begin{equation}
\begin{bmatrix}
x_k\\[1mm]
u_k\\[1mm]
d(x_k,u_k)+\omega_k
\end{bmatrix}
\sim
\mathcal{N}\!\left(
\begin{bmatrix}
\mu_k^x\\[1mm]
\mu_k^u\\[1mm]
\mu_k^d
\end{bmatrix},
\begin{bmatrix}
\Sigma_k^x & \Sigma_k^{xu} & \Sigma_k^{xd}\\
(\Sigma_k^{xu})^\top & \Sigma_k^u & \Sigma_k^{ud}\\
(\Sigma_k^{xd})^\top & (\Sigma_k^{ud})^\top & \Sigma_k^d+\Sigma^\omega
\end{bmatrix}
\right).\label{eqn:normal_dis_all}
\end{equation}
\end{ceqn}
Here, $\mu_k^d$, $\Sigma_k^d$, $\Sigma_k^{xd}$, and $\Sigma_k^{ud}$ are unconditional approximate moments induced by the uncertain query $z_k$. They should not be confused with the deterministic-query GP posterior quantities $\mu^d(z_k)$ and $\Sigma^d(z_k)$ in \eqref{eqn:gp_Bd}. This distinction is important: $\Sigma^d(z_k)$ describes the GP posterior uncertainty at a given input, whereas $\Sigma_k^d$ also includes the additional output variation caused by uncertainty in the input $z_k$.

Using $z_k=[x_k^\top,u_k^\top]^\top$, \eqref{eqn:normal_dis_all} can be written compactly as
\begin{equation}
    \begin{bmatrix}
        z_k\\
        d(z_k)+\omega_k
    \end{bmatrix}
    \sim
    \mathcal{N}\!\left(
    \begin{bmatrix}
        \mu_k^z\\
        \mu_k^d
    \end{bmatrix},
    \begin{bmatrix}
        \Sigma_k^z & \Sigma_k^{zd}\\
        (\Sigma_k^{zd})^\top & \Sigma_k^d+\Sigma^\omega
    \end{bmatrix}
    \right) \, ,\label{eqn:normal_dis_all_compact}
\end{equation}
where
\begin{equation}
    \mu_k^z=\begin{bmatrix}\mu_k^x\\ \mu_k^u\end{bmatrix},\qquad
    \Sigma_k^z=
    \begin{bmatrix}
        \Sigma_k^x & \Sigma_k^{xu}\\
        (\Sigma_k^{xu})^\top & \Sigma_k^u
    \end{bmatrix},\qquad
    \Sigma_k^{zd}=
    \begin{bmatrix}
        \Sigma_k^{xd}\\
        \Sigma_k^{ud}
    \end{bmatrix}.\label{eqn:normal_dis_compact_def}
\end{equation}
The remaining task is to compute the next-state mean $\mu_{k+1}^x$ and covariance $\Sigma_{k+1}^x$ from the moments at time $k$.

\subsection{Mean Propagation Approximation}
\label{sec:mean_prop}
The next-state mean is derived from the law of total expectation. This is the same rule summarized in Appendix~\ref{sec:appendix_gp_approx}, especially \eqref{eqn:mean_prop_general} and its time-indexed form \eqref{eqn:mean_prop_general_final}. Starting from the GP-augmented state-space model in \eqref{eqn:process_model_Bd}, we have
\begin{ceqn}
\begin{align}
    \mu_{k+1}^{x}
    &= m(x_{k+1}) \, ,\label{eqn:mean_approx_general} \\
    &= \mathbb{E}_{z_k}\!\left[
        \mathbb{E}_{d,\omega}\!\left(x_{k+1}\mid z_k\right)
    \right] \, ,\label{eqn:mean_total_expectation_start}\\
    &= \mathbb{E}_{z_k}\!\left[
        \mathbb{E}_{d,\omega}\!\left(
        f(x_k,u_k)+B_d\left(d(x_k,u_k)+\omega_k\right)\mid z_k
        \right)
    \right] \, ,\label{eqn:mean_substitute_model}\\
    &= \mathbb{E}_{z_k}\!\left[
        f(x_k,u_k)+B_d\,\mathbb{E}_{d,\omega}\!\left(d(x_k,u_k)+\omega_k\mid z_k\right)
    \right] \, ,\label{eqn:mean_inner_expectation}\\
    &= \mathbb{E}_{z_k}\!\left[
        f(x_k,u_k)+B_d\,\mu^d(x_k,u_k)
    \right] \, ,\label{eqn:mean_total_expectation}\\
    &= \mathbb{E}_{z_k}\!\left[f(x_k,u_k)\right]
       +\mathbb{E}_{z_k}\!\left[B_d\,\mu^d(x_k,u_k)\right] \, .\label{eqn:mean_split_expanded}
\end{align}
\end{ceqn}
Equation \eqref{eqn:mean_total_expectation_start} applies total expectation: first take the conditional expectation over the GP residual and process noise for a fixed query $z_k$, and then average the result over the uncertain query. In \eqref{eqn:mean_inner_expectation}--\eqref{eqn:mean_total_expectation}, the nominal model $f(x_k,u_k)$ is deterministic once $(x_k,u_k)$ is fixed, $\mathbb{E}[\omega_k]=0$, and the GP posterior mean satisfies $\mathbb{E}[d(x_k,u_k)\mid z_k]=\mu^d(x_k,u_k)$. The two expectations in \eqref{eqn:mean_split_expanded} are still difficult to compute exactly because both $f(x_k,u_k)$ and $\mu^d(x_k,u_k)$ can be nonlinear functions of the Gaussian query.

Using the compact notation $z_k=[x_k^\top,u_k^\top]^\top$, \eqref{eqn:mean_split_expanded} can also be written as
\begin{equation}
    \mu_{k+1}^{x}
    =
    \mathbb{E}_{z_k}\!\left[f(z_k)\right]
    +\mathbb{E}_{z_k}\!\left[B_d\,\mu^d(z_k)\right].\label{eqn:mean_split}
\end{equation}
This compact form is useful for later covariance derivations, but the expanded form in \eqref{eqn:mean_split_expanded} shows explicitly how the state and control input enter the prediction model.

We approximate the nominal model $f(x_k,u_k)$ by a first-order Taylor expansion about the mean point $(\mu_k^x,\mu_k^u)$:
\begin{ceqn}
\begin{align}
    f(x_k,u_k)
    &\approx
    f(\mu_k^x,\mu_k^u)
    +\left.\frac{\partial f(x,u)}{\partial x}\right|_{\mu_k^x,\mu_k^u}(x_k-\mu_k^x)
    +\left.\frac{\partial f(x,u)}{\partial u}\right|_{\mu_k^x,\mu_k^u}(u_k-\mu_k^u) \, ,\label{eqn:nominal_taylor_expanded}\\
    &=
    f(\mu_k^z)+\nabla_z f(\mu_k^z)(z_k-\mu_k^z) \, .\label{eqn:nominal_taylor}
\end{align}
\end{ceqn}
where $\mu_k^z=[(\mu_k^x)^\top,(\mu_k^u)^\top]^\top$ and $\nabla_z f(\mu_k^z)$ is the Jacobian of $f$ with respect to the combined query $z$, evaluated at $\mu_k^z$. Taking the expectation of \eqref{eqn:nominal_taylor_expanded} gives
\begin{ceqn}
\begin{align}
    \mathbb{E}_{z_k}\!\left[f(x_k,u_k)\right]
    &\approx 
    \mathbb{E}_{z_k}\!\left[
    f(\mu_k^x,\mu_k^u)
    +\left.\frac{\partial f(x,u)}{\partial x}\right|_{\mu_k^x,\mu_k^u}(x_k-\mu_k^x)
    +\left.\frac{\partial f(x,u)}{\partial u}\right|_{\mu_k^x,\mu_k^u}(u_k-\mu_k^u)
    \right] \, ,\label{eqn:nominal_mean_expanded}\\
    &=
    f(\mu_k^x,\mu_k^u)
    +\left.\frac{\partial f(x,u)}{\partial x}\right|_{\mu_k^x,\mu_k^u}\mathbb{E}[x_k-\mu_k^x]
    +\left.\frac{\partial f(x,u)}{\partial u}\right|_{\mu_k^x,\mu_k^u}\mathbb{E}[u_k-\mu_k^u] \, ,\label{eqn:nominal_mean_zero_terms}\\
    &=
    f(\mu_k^x,\mu_k^u)
    = f(\mu_k^z).\label{eqn:nominal_mean_taylor_result}
\end{align}
\end{ceqn}
The linear deviation terms vanish because $\mathbb{E}[x_k-\mu_k^x]=0$ and $\mathbb{E}[u_k-\mu_k^u]=0$. This is the GP-MPC version of the generic first-order expectation approximation in \eqref{eqn:exp_mean_appendix}. Higher-order Taylor terms are neglected; therefore, the approximation is most accurate when the uncertainty in $z_k$ is not too large or when the functions are locally close to linear.

Similarly, approximate the GP posterior mean function by its first-order Taylor expansion:
\begin{ceqn}
\begin{align}
    \mu^d(x_k,u_k)
    &\approx
    \mu^d(\mu_k^x,\mu_k^u)
    +\left.\frac{\partial\mu^d(x,u)}{\partial x}\right|_{\mu_k^x,\mu_k^u}(x_k-\mu_k^x)
    +\left.\frac{\partial\mu^d(x,u)}{\partial u}\right|_{\mu_k^x,\mu_k^u}(u_k-\mu_k^u) \, ,\label{eqn:gp_mean_taylor_expanded}\\
    &=
    \mu^d(\mu_k^z)+\nabla_z\mu^d(\mu_k^z)(z_k-\mu_k^z) \, .\label{eqn:gp_mean_taylor}
\end{align}
\end{ceqn}
Thus, the second term in \eqref{eqn:mean_split_expanded} becomes
\begin{ceqn}
\begin{align}
    \mathbb{E}_{z_k}\!\left[B_d\mu^d(x_k,u_k)\right]
    &\approx
    \mathbb{E}_{z_k}\!\left[
    B_d\mu^d(\mu_k^x,\mu_k^u)
    +B_d\left.\frac{\partial\mu^d(x,u)}{\partial x}\right|_{\mu_k^x,\mu_k^u}(x_k-\mu_k^x)
    \right. \nonumber\\
    &\qquad\qquad\left.
    +B_d\left.\frac{\partial\mu^d(x,u)}{\partial u}\right|_{\mu_k^x,\mu_k^u}(u_k-\mu_k^u)
    \right] \, ,\label{eqn:gp_mean_expectation_expanded}\\
    &=
    B_d\mu^d(\mu_k^x,\mu_k^u)
    =B_d\mu^d(\mu_k^z).\label{eqn:gp_mean_taylor_result}
\end{align}
\end{ceqn}
Again, the linear deviation terms vanish under expectation. Substituting \eqref{eqn:nominal_mean_taylor_result} and \eqref{eqn:gp_mean_taylor_result} into \eqref{eqn:mean_split_expanded} gives the propagated mean approximation:
\begin{equation}
    \mu_{k+1}^{x}
    \approx
    f(\mu_k^x,\mu_k^u)+B_d\mu^d(\mu_k^x,\mu_k^u)
    =
    f(\mu_k^z)+B_d\mu^d(\mu_k^z) \, .\label{eqn:mean_prop_final_again}
\end{equation}
This equation is the mean prediction model used in both implementation families discussed later: mean-only GP predictive control and uncertainty-aware constrained GP-MPC. The difference between these families lies not in the propagated mean, but in whether and how the covariance $\Sigma_{k+1}^x$ is propagated and used.

\subsection{Uncertainty Propagation Approximation}
\label{sec:uncertainty_prop}
\label{sec:var_prop}

\subsubsection{Taylor Approximation}
Uncertainty propagation is essential when the controller needs to reason about risk, chance constraints, robustness margins, or confidence in future predictions. The Taylor approximation is one widely used approach because it provides a tractable first-order covariance recursion. It propagates uncertainty through the local sensitivities of the nominal model and the GP posterior mean, while also retaining the GP posterior covariance evaluated at the query mean.

The covariance propagation follows from the law of total covariance, given in generic form in \eqref{eqn:var_prop_general} and in time-indexed form in \eqref{eqn:var_prop_general_final}. Starting again from \eqref{eqn:process_model_Bd}, we obtain
\begin{ceqn}
\begin{align}
    \Sigma_{k+1}^{x}
    &=\operatorname{var}(x_{k+1}) \, ,\label{eqn:var_prop_general_intro}\\
    &=
    \mathbb{E}_{z_k}\!\left[
        \operatorname{var}_{d,\omega}\!\left(x_{k+1}\mid z_k\right)
    \right]
    +
    \operatorname{var}_{z_k}\!\left(
        \mathbb{E}_{d,\omega}\!\left[x_{k+1}\mid z_k\right]
    \right) \, ,\label{eqn:state_var_total_covariance}\\
    &=
    \mathbb{E}_{z_k}\!\left[
        B_d\left(\Sigma^d(z_k)+\Sigma^\omega\right)B_d^\top
    \right]
    +
    \operatorname{var}_{z_k}\!\left(
        f(x_k,u_k)+B_d\mu^d(x_k,u_k)
    \right) \, ,\label{eqn:var_prop_law_expanded}\\
    &=
    \mathbb{E}_{z_k}\!\left[
        B_d\left(\Sigma^d(z_k)+\Sigma^\omega\right)B_d^\top
    \right]
    +
    \operatorname{var}_{z_k}\!\left(
        f(z_k)+B_d\mu^d(z_k)
    \right) \, .\label{eqn:var_prop_law}
\end{align}
\end{ceqn}
The first term in \eqref{eqn:var_prop_law} is the expected conditional covariance. It comes from the GP posterior covariance and the process noise when the query is fixed. The second term is the covariance of the conditional mean. It appears because the query $z_k$ is uncertain and is passed through the nominal model and the GP posterior mean.

Define the Jacobians
\begin{equation*}
    F_k \triangleq \nabla_z f(\mu_k^z),
    \qquad
    D_k \triangleq \nabla_z\mu^d(\mu_k^z).
\end{equation*}
Using the first-order approximations \eqref{eqn:nominal_taylor} and \eqref{eqn:gp_mean_taylor}, together with \eqref{eqn:exp_var_appendix} and \eqref{eqn:var_mean_appendix}, the two terms in \eqref{eqn:var_prop_law} can be approximated as follows:
\begin{ceqn}
\begin{align}
    \mathbb{E}_{z_k}\!\left[
        B_d\left(\Sigma^d(z_k)+\Sigma^\omega\right)B_d^\top
    \right]
    &\approx
    B_d\left(\Sigma^d(\mu_k^z)+\Sigma^\omega\right)B_d^\top \, ,\label{eqn:expected_conditional_cov}\\
    \operatorname{var}_{z_k}\!\left(
        f(z_k)+B_d\mu^d(z_k)
    \right)
    &\approx
    \operatorname{var}_{z_k}\!\left(
        f(\mu_k^z)+F_k(z_k-\mu_k^z)
        +B_d\mu^d(\mu_k^z)+B_dD_k(z_k-\mu_k^z)
    \right) \, ,\label{eqn:variance_linearized_mean}\\
    &=
    \operatorname{var}_{z_k}\!\left(
        (F_k+B_dD_k)(z_k-\mu_k^z)
    \right) \, ,\label{eqn:variance_linearized_mean_reduced}\\
    &=
    (F_k+B_dD_k)\Sigma_k^z(F_k+B_dD_k)^\top .\label{eqn:variance_linearized_mean_final}
\end{align}
\end{ceqn}
The constant terms $f(\mu_k^z)$ and $B_d\mu^d(\mu_k^z)$ disappear in the variance calculation because constants have zero variance. Substituting \eqref{eqn:expected_conditional_cov} and \eqref{eqn:variance_linearized_mean_final} into \eqref{eqn:var_prop_law} yields
\begin{ceqn}
\begin{align}
    \Sigma_{k+1}^{x}
    &\approx
    B_d\left(\Sigma^d(\mu_k^z)+\Sigma^\omega\right)B_d^\top
    +(F_k+B_dD_k)\Sigma_k^z(F_k+B_dD_k)^\top \, ,\label{eqn:var_prop_taylor_compact}\\
    &=
    B_d\left(\Sigma^d(\mu_k^z)+\Sigma^\omega\right)B_d^\top
    +F_k\Sigma_k^zF_k^\top
    +F_k\Sigma_k^zD_k^\top B_d^\top
    \nonumber\\
    &\qquad
    +B_dD_k\Sigma_k^zF_k^\top
    +B_dD_k\Sigma_k^zD_k^\top B_d^\top .\label{eqn:var_prop_taylor_expanded}
\end{align}
\end{ceqn}
The expanded expression \eqref{eqn:var_prop_taylor_expanded} is useful pedagogically because each term has a clear role. The term $F_k\Sigma_k^zF_k^\top$ is the query uncertainty propagated through the nominal model. The term $B_dD_k\Sigma_k^zD_k^\top B_d^\top$ is the additional uncertainty induced by passing the uncertain query through the GP posterior mean. The two cross terms $F_k\Sigma_k^zD_k^\top B_d^\top$ and $B_dD_k\Sigma_k^zF_k^\top$ capture the correlation between the nominal-model variation and the learned residual variation. Finally, $B_d(\Sigma^d(\mu_k^z)+\Sigma^\omega)B_d^\top$ represents the GP posterior covariance and process-noise contribution.

The same result can be rewritten in the joint-covariance form used in \eqref{eqn:normal_dis_all_compact}:
\begin{ceqn}
\begin{align}
    \Sigma_{k+1}^{x}
    &\approx
    \begin{bmatrix}F_k & B_d\end{bmatrix}
    \begin{bmatrix}
        \Sigma_k^z & \Sigma_k^zD_k^\top\\
        D_k\Sigma_k^z & \Sigma^d(\mu_k^z)+D_k\Sigma_k^zD_k^\top+\Sigma^\omega
    \end{bmatrix}
    \begin{bmatrix}F_k & B_d\end{bmatrix}^{\top} \, ,\label{eqn:var_prop_joint_first}\\
    &=
    \begin{bmatrix}F_k & B_d\end{bmatrix}
    \begin{bmatrix}
        \Sigma_k^z & \Sigma_k^{zd}\\
        (\Sigma_k^{zd})^\top & \Sigma_k^d+\Sigma^\omega
    \end{bmatrix}
    \begin{bmatrix}F_k & B_d\end{bmatrix}^{\top} \, ,\label{eqn:var_prop_joint_second}\\
    &=
    \begin{bmatrix}F_k & B_d\end{bmatrix}
    \Sigma_k
    \begin{bmatrix}F_k & B_d\end{bmatrix}^{\top}.\label{eqn:var_prop_taylor}
\end{align}
\end{ceqn}
Comparing \eqref{eqn:var_prop_joint_first} and \eqref{eqn:var_prop_joint_second}, the Taylor approximation gives
\begin{equation}
    \begin{bmatrix}
        \Sigma_k^{zd}\\
        \Sigma_k^d
    \end{bmatrix}
    \approx
    \begin{bmatrix}
        \Sigma_k^zD_k^\top\\
        \Sigma^d(\mu_k^z)+D_k\Sigma_k^zD_k^\top
    \end{bmatrix}.\label{eqn:covariance_terms_taylor}
\end{equation}
Equation \eqref{eqn:covariance_terms_taylor} shows that the unconditional GP output covariance $\Sigma_k^d$ contains two components: the deterministic-query GP posterior covariance $\Sigma^d(\mu_k^z)$ and the additional covariance $D_k\Sigma_k^zD_k^\top$ caused by uncertainty in the GP input. It also shows that the query--GP output cross-covariance $\Sigma_k^{zd}$ is generally nonzero when the GP posterior mean changes with the query. Omitting these terms simplifies computation, but it also changes the uncertainty model used by the MPC controller.

\subsubsection{Mean-Equivalent Approximation}
The mean-equivalent approximation is a simpler and lower-cost covariance propagation strategy \cite{langaaker2018cautious}. Instead of propagating input uncertainty through the GP posterior mean, it evaluates GP quantities at the query mean and neglects the cross-covariance between the query and the GP output:
\begin{equation}
    \begin{bmatrix}
        \Sigma_k^{zd}\\
        \Sigma_k^d
    \end{bmatrix}
    \approx
    \begin{bmatrix}
        0\\
        \Sigma^d(\mu_k^z)
    \end{bmatrix}.\label{eqn:mean_equiv_cov_terms}
\end{equation}
This approximation keeps the GP posterior covariance at the mean input, but it omits the terms $\Sigma_k^zD_k^\top$ and $D_k\Sigma_k^zD_k^\top$ from \eqref{eqn:covariance_terms_taylor}. Substituting \eqref{eqn:mean_equiv_cov_terms} into the joint covariance form gives
\begin{ceqn}
\begin{align}
    \Sigma_{k+1}^{x}
    &\approx
    \begin{bmatrix}F_k & B_d\end{bmatrix}
    \begin{bmatrix}
        \Sigma_k^z & 0\\
        0 & \Sigma^d(\mu_k^z)+\Sigma^\omega
    \end{bmatrix}
    \begin{bmatrix}F_k & B_d\end{bmatrix}^{\top} \, ,\label{eqn:var_prop_mean_equiv_joint}\\
    &=
    F_k\Sigma_k^zF_k^\top
    +
    B_d\left(\Sigma^d(\mu_k^z)+\Sigma^\omega\right)B_d^\top .\label{eqn:var_prop_mean_equiv}
\end{align}
\end{ceqn}
The mean-equivalent approximation is computationally attractive because it avoids computing and propagating the GP mean Jacobian contribution to the output covariance. However, it is less complete than the Taylor approximation because it does not account for how uncertainty in $z_k$ changes the GP residual mean. It can therefore underestimate or misrepresent uncertainty when the GP posterior mean varies strongly over the region covered by the predicted state distribution. The mixed-platoon example in Sec.~\ref{sec:gp_mixed_platoon} uses this type of simplified covariance propagation for the human-driven vehicle position variance, and the neglected cross-covariances are stated explicitly there.

\subsection{Summary of Mean and Uncertainty Approximation}
\label{sec:mean_var_summary}
The derivation above leads to the following general one-step moment propagation equations for the GP-augmented model \eqref{eqn:process_model_Bd}:
\begin{ceqn}
\begin{align}
    \mu_{k+1}^{x}
    &\approx f(\mu_k^z)+B_d\mu^d(\mu_k^z),\label{eqn:mean_summary}\\
    \Sigma_{k+1}^{x}
    &\approx
    \begin{bmatrix}F_k & B_d\end{bmatrix}
    \Sigma_k
    \begin{bmatrix}F_k & B_d\end{bmatrix}^{\top}.\label{eqn:var_summary}
\end{align}
\end{ceqn}
The mean equation \eqref{eqn:mean_summary} is obtained by evaluating the nominal model and the GP posterior mean at the query mean. The covariance equation \eqref{eqn:var_summary} has the same outer structure for both Taylor and mean-equivalent propagation; the difference lies in how the joint covariance matrix $\Sigma_k$ is constructed. The Taylor method uses \eqref{eqn:covariance_terms_taylor}, whereas the mean-equivalent method uses \eqref{eqn:mean_equiv_cov_terms}. Thus, the key modeling choice is whether to retain or neglect the GP input-uncertainty terms $\Sigma_k^zD_k^\top$ and $D_k\Sigma_k^zD_k^\top$.

This distinction gives a practical trade-off between computational efficiency and uncertainty fidelity. The Taylor approximation includes first-order sensitivity of the GP posterior mean and is therefore more informative when the predicted query distribution covers a region where the learned residual changes significantly. The mean-equivalent approximation is simpler and faster, but it treats the GP model as if it were evaluated only at the mean query for covariance purposes. For real-time MPC, the appropriate choice depends on the prediction horizon, state dimension, GP output dimension, available computation time, and whether the propagated covariance is used only for monitoring or directly inside chance constraints.

Beyond these two approximations, exact moment matching \cite{deisenroth2010efficient} computes the GP predictive mean and variance by integrating over the Gaussian input distribution rather than using a first-order linearization. This can provide a more accurate single-step moment approximation under compatible kernels, but it is typically more computationally demanding. Moreover, in multi-step MPC the state distribution is repeatedly approximated as Gaussian even though the true distribution can become non-Gaussian after passing through nonlinear GP-augmented dynamics. As discussed in \cite{hewing2018cautious}, the practical benefit of exact moment matching over Taylor approximation can therefore be limited in typical GP-MPC prediction loops, especially when real-time solution speed is important.

In summary, these methods form a spectrum. Mean-only prediction uses the GP posterior mean but does not propagate covariance. Mean-equivalent covariance propagation keeps the GP posterior covariance at the mean query but neglects GP input-uncertainty effects. Taylor covariance propagation retains first-order input-uncertainty effects through the GP mean Jacobian. Exact moment matching aims for more accurate moment integration but usually increases computational cost. Table~\ref{tab:propagation_methods} summarizes how these options are used or discussed in this tutorial.
\begin{table}[t]
\caption{Mean and uncertainty propagation options used or discussed in this tutorial.}
\label{tab:propagation_methods}
\centering
\footnotesize
\begin{tabular}{@{}p{0.24\linewidth}p{0.12\linewidth}p{0.12\linewidth}p{0.21\linewidth}p{0.19\linewidth}@{}}
\toprule
Method & GP mean & GP variance & Uncertain-input contribution & Role in this tutorial \\
\midrule
Mean-only prediction & Yes & No & No covariance propagation & Path-following examples \\
Mean-equivalent covariance & Yes & Yes & GP cross-covariance neglected & Mixed-platoon example \\
Taylor covariance & Yes & Yes & First-order sensitivity through $D_k$ & Main step-by-step derivation \\
Moment matching & Yes & Yes & Moment integration under Gaussian input & Alternative discussed \\
\bottomrule
\end{tabular}
\end{table}

\subsection{Mean and Uncertainty Approximation with Linear Nominal Model}
\label{sec:linear_nominal}
The preceding derivation applies to a general nonlinear nominal model. Many GP-MPC formulations, however, use a linear nominal model or a locally linearized nominal model. This special case is useful because it shows how the compact covariance formula reduces to the matrix expressions commonly used in cautious and chance-constrained GP-MPC.

Assume that the nominal model is linear,
\begin{equation*}
    f(x_k,u_k)=Ax_k+Bu_k,
\end{equation*}
where $A$ and $B$ are the state and input matrices. Then the nominal-model Jacobian is
\begin{equation*}
    F_k=\nabla_z f(\mu_k^z)=\begin{bmatrix}A & B\end{bmatrix}.
\end{equation*}
Substituting this into \eqref{eqn:var_summary} and using the expanded joint covariance in \eqref{eqn:normal_dis_all} gives
\begin{ceqn}
\begin{align}
    \Sigma_{k+1}^{x}
    &=
    \begin{bmatrix}A & B & B_d\end{bmatrix}
    \begin{bmatrix}
        \Sigma_k^x & \Sigma_k^{xu} & \Sigma_k^{xd}\\
        (\Sigma_k^{xu})^\top & \Sigma_k^u & \Sigma_k^{ud}\\
        (\Sigma_k^{xd})^\top & (\Sigma_k^{ud})^\top & \Sigma_k^d+\Sigma^\omega
    \end{bmatrix}
    \begin{bmatrix}A & B & B_d\end{bmatrix}^{\top}.\label{eqn:linear_cov_general}
\end{align}
\end{ceqn}
Equation \eqref{eqn:linear_cov_general} is simply the covariance propagation of the linear combination $Ax_k+Bu_k+B_d(d_k+\omega_k)$ using the approximate joint Gaussian moments in \eqref{eqn:normal_dis_all}. It makes explicit that the covariance of the next state depends not only on the state and input covariances, but also on the covariance of the learned residual and its cross-covariance with the state and input.

For the mean-equivalent approximation, the GP-output cross-covariances are neglected and the GP-output covariance is evaluated at the query mean. Thus, \eqref{eqn:linear_cov_general} becomes
\begin{ceqn}
\begin{align}
    \Sigma_{k+1}^{x}
    &\approx
    \begin{bmatrix}A & B & B_d\end{bmatrix}
    \begin{bmatrix}
        \Sigma_k^x & \Sigma_k^{xu} & 0\\
        (\Sigma_k^{xu})^\top & \Sigma_k^u & 0\\
        0 & 0 & \Sigma^d(\mu_k^z)+\Sigma^\omega
    \end{bmatrix}
    \begin{bmatrix}A & B & B_d\end{bmatrix}^{\top}.\label{eqn:linear_cov_mean_equiv}
\end{align}
\end{ceqn}
For the Taylor approximation, write the GP mean Jacobian as
\begin{equation*}
    D_k=\nabla_z\mu^d(\mu_k^z)=\begin{bmatrix}D_k^x & D_k^u\end{bmatrix},
\end{equation*}
where $D_k^x=\left.\partial\mu^d(x,u)/\partial x\right|_{\mu_k^x,\mu_k^u}$ and $D_k^u=\left.\partial\mu^d(x,u)/\partial u\right|_{\mu_k^x,\mu_k^u}$. Since
\begin{equation*}
    \Sigma_k^zD_k^\top
    =
    \begin{bmatrix}
        \Sigma_k^x & \Sigma_k^{xu}\\
        (\Sigma_k^{xu})^\top & \Sigma_k^u
    \end{bmatrix}
    \begin{bmatrix}
        (D_k^x)^\top\\
        (D_k^u)^\top
    \end{bmatrix},
\end{equation*}
the Taylor cross-covariance terms become
\begin{align*}
    \Sigma_k^{xd}
    &\approx
    \Sigma_k^x(D_k^x)^\top+\Sigma_k^{xu}(D_k^u)^\top,\\
    \Sigma_k^{ud}
    &\approx
    (\Sigma_k^{xu})^\top(D_k^x)^\top+\Sigma_k^u(D_k^u)^\top.
\end{align*}
Similarly, the GP-output covariance contribution induced by uncertain inputs is
\begin{align*}
    D_k\Sigma_k^zD_k^\top
    &=
    \begin{bmatrix}D_k^x & D_k^u\end{bmatrix}
    \begin{bmatrix}
        \Sigma_k^x & \Sigma_k^{xu}\\
        (\Sigma_k^{xu})^\top & \Sigma_k^u
    \end{bmatrix}
    \begin{bmatrix}
        (D_k^x)^\top\\
        (D_k^u)^\top
    \end{bmatrix}\\
    &=
    D_k^x\Sigma_k^x(D_k^x)^\top
    +D_k^x\Sigma_k^{xu}(D_k^u)^\top
    +D_k^u(\Sigma_k^{xu})^\top(D_k^x)^\top
    +D_k^u\Sigma_k^u(D_k^u)^\top .
\end{align*}
Combining these terms gives the Taylor covariance components
\begin{ceqn}
\begin{align}
    \Sigma_k^{xd}
    &\approx
    \Sigma_k^x(D_k^x)^\top+\Sigma_k^{xu}(D_k^u)^\top, \nonumber\\
    \Sigma_k^{ud}
    &\approx
    (\Sigma_k^{xu})^\top(D_k^x)^\top+\Sigma_k^u(D_k^u)^\top, \nonumber\\
    \Sigma_k^d
    &\approx
    \Sigma^d(\mu_k^z)
    +D_k^x\Sigma_k^x(D_k^x)^\top
    +D_k^x\Sigma_k^{xu}(D_k^u)^\top
    +D_k^u(\Sigma_k^{xu})^\top(D_k^x)^\top
    +D_k^u\Sigma_k^u(D_k^u)^\top .\label{eqn:linear_cov_taylor_components}
\end{align}
\end{ceqn}
Equations \eqref{eqn:linear_cov_general}--\eqref{eqn:linear_cov_taylor_components} recover the linear-nominal-model case commonly used in cautious and chance-constrained GP-MPC formulations. They also show which terms disappear in standard deterministic-input open-loop MPC prediction. If the future input sequence is deterministic during the optimization evaluation, then $\Sigma_k^u=0$ and $\Sigma_k^{xu}=0$; the remaining uncertainty comes from the predicted state covariance, the GP posterior covariance, and, under the Taylor approximation, the sensitivity of the GP posterior mean to the uncertain state.

\section{GP-MPC Problem Formulations and Implementation Workflow}
\label{sec:gp_mpc}
This section defines the control problems to which the GP predictions in Sec.~\ref{sec:mean_var_approx} are applied. A central distinction is whether the controller uses only the GP posterior mean to improve the prediction model or also propagates GP uncertainty to impose probabilistic constraints. The former gives a mean-only unconstrained GP predictive controller when no inequality constraints are included; the latter gives an uncertainty-aware constrained GP-MPC formulation. Both are receding-horizon controllers, but they support different claims regarding constraint handling and safety.

\subsection{Regulation and Output-Tracking Problems}
\label{sec:regulation_tracking}

Before introducing specific GP-MPC formulations, it is useful to distinguish two common control objectives: regulation and output tracking. Let the GP-augmented state prediction model be given by \eqref{eqn:process_model_Bd}. The state is denoted by $x_k\in\mathbb{R}^{n_x}$, the control input by $u_k\in\mathbb{R}^{n_u}$, and the controlled output by
\[
y_k=h(x_k)\in\mathbb{R}^{n_y} .
\]
Here, $h(\cdot)$ is an output map that selects or transforms the state variables used in the control objective. For example, $h(x)=x$ corresponds to full-state control, while in a mobile-robot path-following problem $h(x)$ may represent path-error coordinates rather than the full robot pose. At time $k$, predictions over a horizon of length $N$ use the notation $x_{i|k}$, $u_{i|k}$, and $y_{i|k}$ for $i=0,\ldots,N$, where the subscript $i|k$ means ``the value predicted $i$ steps ahead using information available at time $k$.''

A \emph{regulation} problem drives the system toward a feasible fixed point $(x_r,u_r)$. For the learned mean model, this fixed point should satisfy
\begin{equation}
x_r=f(x_r,u_r)+B_d\mu^d(x_r,u_r) \, .\label{eqn:feasible_equilibrium}
\end{equation}
A reference-compatible regulation cost should therefore penalize deviations $x_{i|k}-x_r$ and $u_{i|k}-u_r$. Penalizing the absolute input $u_{i|k}$ toward zero is only appropriate when the equilibrium input is actually $u_r=0$; otherwise, it may bias the optimizer toward an input that is incompatible with the desired equilibrium and can introduce steady-state offset.

An \emph{output-tracking} problem instead drives selected outputs toward a desired reference trajectory $r_{i|k}$. In this case, the objective is not necessarily to make the full state equal to a fixed point, but to make the controlled output $y_{i|k}=h(x_{i|k})$ follow the reference as well as the system dynamics, actuation limits, and constraints allow. A general quadratic output-tracking cost is
\begin{align}
J_k(\mathbf U_k) =
\sum_{i=0}^{N-1}\left(
\lVert \bar y_{i|k}-r_{i|k}\rVert_Q^2
+\lVert u_{i|k}-u_{r,i|k}\rVert_R^2
+\lVert \Delta u_{i|k}\rVert_S^2\right) 
+\lVert \bar y_{N|k}-r_{N|k}\rVert_{Q_f}^2 .\label{eqn:reference_cost}
\end{align}
Here, $\mathbf U_k=(u_{0|k},\ldots,u_{N-1|k})$ is the sequence of control inputs optimized at time $k$, and $\Delta u_{i|k}=u_{i|k}-u_{i-1|k}$ denotes the change in control input between two consecutive prediction steps. The notation $\lVert e\rVert_Q^2$ denotes a weighted squared error, so that larger entries in the weighting matrix assign larger penalties to the corresponding components of the error. Specifically, $Q$ penalizes the output-tracking error $\bar y_{i|k}-r_{i|k}$ along the prediction horizon, $R$ penalizes the deviation of the control input from the reference input $u_{r,i|k}$, $S$ penalizes rapid changes in the control input, and $Q_f$ penalizes the terminal output-tracking error at the end of the horizon. These weighting matrices are typically chosen to be positive semidefinite.

The reference inputs $u_{r,i|k}$ should be compatible with the desired trajectory whenever they are known. This compatibility is important because exact independent tracking of arbitrary output components also requires a dynamically feasible reference and sufficient actuation. In general, one cannot require arbitrary exact tracking of more independent outputs than the available control degrees of freedom can support. For this reason, the mobile-robot example in Sec.~\ref{sec:path_following} is stated as a path-error output-tracking problem, rather than arbitrary independent tracking of all pose coordinates.

\subsection{Mean-Only Unconstrained GP Predictive Control}
\label{sec:mean_only_mpc}

The least computationally demanding use of a GP within receding-horizon control is to include only its posterior mean in the prediction model. In this formulation, the GP acts as a deterministic correction to the physics-based nominal model, while the GP posterior covariance is not propagated inside the online optimization problem. The result is a simple and efficient GP-MPC formulation that is useful when the main objective is to compensate for model mismatch, but not to explicitly enforce probabilistic constraints.

Let $\hat x_k$ denote the measured or estimated state available at time $k$. Starting from this state, the controller predicts a mean state trajectory over the horizon. The notation $\bar x_{i|k}$ denotes the predicted mean state at prediction step $i$, computed at time $k$, and $\bar y_{i|k}=h(\bar x_{i|k})$ denotes the corresponding predicted mean output. 

With no state or input inequality constraints, the mean-only output-tracking problem is written as
\begin{align}
    \min_{\mathbf U_k}\quad
        & J_k(\mathbf U_k)\label{eqn:mean_only_objective}\\
    \text{s.t.}\quad
        & \bar x_{0|k}=\hat x_k\label{eqn:mean_only_initial}\\
        & \bar x_{i+1|k}
        = f(\bar x_{i|k},u_{i|k})
        +B_d\mu^d(\bar x_{i|k},u_{i|k}),
        \quad i=0,\ldots,N-1\label{eqn:mean_only_dynamics}\\
        & \bar y_{i|k}=h(\bar x_{i|k}),
        \quad i=0,\ldots,N .\label{eqn:mean_only_output}
\end{align}
Here, $\mathbf U_k=(u_{0|k},\ldots,u_{N-1|k})$ is the sequence of control inputs optimized at time $k$, and $J_k(\mathbf U_k)$ is the tracking cost defined in \eqref{eqn:reference_cost}. The function $f(\cdot,\cdot)$ is the physics-based nominal model, $\mu^d(\cdot,\cdot)$ is the GP posterior mean of the learned residual dynamics, and $B_d$ maps the learned residual into the state equation. Equation~\eqref{eqn:mean_only_initial} initializes the prediction from the current state estimate. Equation~\eqref{eqn:mean_only_dynamics} then predicts the next mean state by adding the nominal model prediction and the GP mean correction. Finally, \eqref{eqn:mean_only_output} converts each predicted mean state into the output variables that appear in the tracking cost.

This formulation is called ``mean-only'' because the online optimizer uses the GP posterior mean $\mu^d$ but does not directly use the GP posterior covariance $\Sigma^d$ or the propagated state covariance $\Sigma^x$. Therefore, it does not explicitly account for uncertainty growth along the prediction horizon. Its main advantage is computational efficiency: the optimization problem has essentially the same structure as a deterministic nonlinear MPC problem, with the GP mean acting as an additional learned model term. Its main limitation is that uncertainty information cannot be used directly to tighten constraints, penalize prediction variance, or enforce chance constraints.

The mean-only formulation describes the role of GP means in the path-following examples in Sec.~\ref{sec:path_following}. In those examples, the learned GP term improves the predicted path-error dynamics, while the controller optimizes a deterministic mean trajectory. Where the published implementation uses a zero reference for a virtual error-regulating input, the absolute-input term in its reported quadratic cost is the special case $u_{r,i|k}=0$ of \eqref{eqn:reference_cost}.

\subsection{Uncertainty-Aware Constrained GP-MPC}
\label{sec:constrained_gp_mpc}

The mean-only formulation in Sec.~\ref{sec:mean_only_mpc} improves the prediction model by using the GP posterior mean, but it does not use the GP posterior covariance during online optimization. When state constraints, safety margins, or collision-avoidance requirements are central to the task, this may be insufficient because prediction uncertainty can grow along the horizon. In such cases, the controller propagates both the predicted state mean and covariance and uses the covariance to tighten probabilistic constraints.

Let $\mu_{i|k}^x$ and $\Sigma_{i|k}^x$ denote the predicted mean and covariance of the state $x_{i|k}$ at prediction step $i$, computed at time $k$. The initial mean is the current state estimate $\hat x_k$, and the initial covariance $\hat\Sigma_k^x$ represents the uncertainty of this estimate. For example, $\hat\Sigma_k^x$ may come from a state estimator, or it may be set to a small positive semidefinite matrix when the current state is assumed to be accurately measured.

The GP residual model is evaluated using the query vector $z_{i|k}$, which combines the predicted state and input. In the open-loop MPC formulation used here, future inputs are deterministic decision variables, while future states are uncertain because model uncertainty and prediction uncertainty are propagated over the horizon. Therefore,
\begin{equation}
    \mu_{i|k}^z=
    \begin{bmatrix}
        \mu_{i|k}^x\\
        u_{i|k}
    \end{bmatrix},
    \qquad
    \Sigma_{i|k}^z=
    \begin{bmatrix}
        \Sigma_{i|k}^x & 0\\
        0 & 0
    \end{bmatrix}.\label{eqn:query_mean_cov_constrained}
\end{equation}
The zero blocks in $\Sigma_{i|k}^z$ indicate that the input sequence is optimized deterministically in this open-loop prediction. If a feedback policy or uncertain future inputs are used, the input-related blocks of $\Sigma_{i|k}^z$ can be modified accordingly.

A representative uncertainty-aware constrained GP-MPC formulation is
\begin{align}
    \min_{\mathbf U_k}\quad
        & J_k(\mathbf U_k)\label{eqn:constrained_objective}\\
    \text{s.t.}\quad
        & \mu_{0|k}^x=\hat x_k,\qquad
          \Sigma_{0|k}^x=\hat\Sigma_k^x\label{eqn:constrained_initial}\\
        & \mu_{i+1|k}^x
          = f(\mu_{i|k}^x,u_{i|k})
          +B_d\mu^d(\mu_{i|k}^z),
          \quad i=0,\ldots,N-1\label{eqn:constrained_mean}\\
        & \Sigma_{i+1|k}^x
          =\mathcal{P}(\mu_{i|k}^z,\Sigma_{i|k}^z),
          \quad i=0,\ldots,N-1\label{eqn:constrained_cov}\\
        & u_{i|k}\in\mathcal U,
          \quad i=0,\ldots,N-1\label{eqn:constrained_input}\\
        & \operatorname{Pr}\!\left(a_j^\top x_{i|k}\leq b_j\right)\geq p_j,
          \quad j=1,\ldots,n_c,\quad i=1,\ldots,N .\label{eqn:constrained_chance}
\end{align}
Here, $\mathbf U_k=(u_{0|k},\ldots,u_{N-1|k})$ is the optimized input sequence, and $J_k(\mathbf U_k)$ can be chosen as the tracking cost in \eqref{eqn:reference_cost}. Equation~\eqref{eqn:constrained_initial} initializes both the predicted state mean and covariance. Equation~\eqref{eqn:constrained_mean} propagates the predicted mean using the physics-based nominal model $f(\cdot,\cdot)$ evaluated at the predicted mean state and deterministic input, together with the GP posterior mean $\mu^d(\cdot)$ of the learned residual dynamics evaluated at the corresponding query. Equation~\eqref{eqn:constrained_cov} propagates the predicted covariance using a selected approximation method $\mathcal{P}(\cdot)$, such as the Taylor approximation in \eqref{eqn:var_prop_taylor} or the mean-equivalent approximation in \eqref{eqn:var_prop_mean_equiv}. The set $\mathcal U$ denotes the admissible input set, such as actuator saturation limits. Since the inputs are deterministic decision variables in this open-loop formulation, input bounds are imposed directly as hard constraints.

The final constraint in \eqref{eqn:constrained_chance} is a linear chance constraint on the uncertain predicted state. The vector $a_j$ selects the constrained direction or state combination, $b_j$ is the corresponding constraint bound, $p_j\in(0,1)$ is the required probability of satisfying the constraint, and $n_c$ is the number of scalar chance constraints. For example, a constraint on vehicle spacing, obstacle clearance, or position error can often be written or locally approximated in the form $a_j^\top x_{i|k}\leq b_j$. A larger value of $p_j$ gives a more conservative constraint because it requires a higher probability of satisfaction.

Under the Gaussian approximation $x_{i|k}\sim\mathcal N(\mu_{i|k}^x,\Sigma_{i|k}^x)$, the scalar random variable $a_j^\top x_{i|k}$ is also Gaussian, with mean $a_j^\top\mu_{i|k}^x$ and variance $a_j^\top\Sigma_{i|k}^x a_j$. Therefore, the chance constraint in \eqref{eqn:constrained_chance} can be converted into the deterministic tightened constraint
\begin{equation}
    a_j^\top\mu_{i|k}^x
    +\Phi^{-1}(p_j)
    \sqrt{a_j^\top\Sigma_{i|k}^x a_j}
    \leq b_j \, .\label{eqn:generic_tightening}
\end{equation}
where $\Phi^{-1}$ is the inverse cumulative distribution function of the standard normal distribution. The first term, $a_j^\top\mu_{i|k}^x$, is the nominal predicted value of the constrained quantity. The second term is an uncertainty margin. It increases when the predicted covariance in the constrained direction is larger or when the required satisfaction probability $p_j$ is closer to one. Thus, uncertainty propagation affects the controller by shrinking the feasible region available to the predicted mean trajectory.

This formulation makes explicit how GP uncertainty enters constrained MPC. The GP posterior mean improves the predicted dynamics, while the propagated covariance determines how much each probabilistic constraint should be tightened. The mixed-vehicle platooning controller in Sec.~\ref{sec:gp_mixed_platoon} is an explicit constrained GP-MPC implementation: it uses a GP-based human-vehicle model, propagates position uncertainty using a stated mean-equivalent approximation, and tightens the minimum safe-distance constraint according to \eqref{eqn:generic_tightening}.

\begin{figure}[t]
\centering
\begin{tikzpicture}[
    node distance=2.8mm and 3.0mm,
    every node/.style={font=\scriptsize, align=center},
    block/.style={draw, rounded corners, minimum height=6.0mm, text width=0.25\linewidth, inner sep=2pt},
    decision/.style={draw, rounded corners, minimum height=6.0mm, text width=0.30\linewidth, inner sep=2pt},
    arrow/.style={-{Latex[length=2mm]}, thin}
]
\node[block] (data) {Offline: nominal model; residual data};
\node[block, right=of data] (gp) {Train GP or sparse GP; validate calibration};
\node[block, right=of gp] (design) {Select horizon, weights, constraints, risk level};
\draw[arrow] (data) -- (gp);
\draw[arrow] (gp) -- (design);

\node[block, below=7mm of gp] (state) {Online: state estimate; reference horizon};
\node[decision, below=of state] (pred) {GP mean prediction; optional covariance propagation};
\node[block, below left=4mm and -4mm of pred] (mean) {Mean-only unconstrained solve};
\node[block, below right=4mm and -4mm of pred] (cons) {Tighten constraints; constrained solve};
\node[block, below=11mm of pred] (apply) {Apply first input; shift horizon; repeat};

\draw[arrow] (design) |- (state);
\draw[arrow] (state) -- (pred);
\draw[arrow] (pred) -- node[left, xshift=-1mm] {mean} (mean);
\draw[arrow] (pred) -- node[right, xshift=1mm] {mean + variance} (cons);
\draw[arrow] (mean) -- (apply);
\draw[arrow] (cons) -- (apply);
\end{tikzpicture}
\caption{General GP-MPC workflow. The path-following examples use the mean-only branch, whereas the mixed-platoon example uses propagated uncertainty and tightened constraints.}
    \Description{A flowchart showing offline GP-MPC preparation and the two online branches: mean-only unconstrained control and uncertainty-aware constrained control.}
    \label{fig:gp_mpc_workflow}
\end{figure}

\subsection{Linearized Mean-Only Prediction in Lifted Form}
\label{sec:mean_only_lifted}

The mean-only GP-MPC problem in Sec.~\ref{sec:mean_only_mpc} is generally nonlinear because both the physics-based nominal model $f(\cdot,\cdot)$ and the GP posterior mean $\mu^d(\cdot,\cdot)$ may be nonlinear functions of the state and input. For efficient unconstrained implementation, a common approach is to linearize the mean-only prediction model along a nominal trajectory and then write the predicted deviations over the whole horizon in a lifted matrix form. This gives a local quadratic approximation of the MPC problem and leads to an analytical input update in the unconstrained case.

Let $\bar x_{i|k}$ and $\bar u_{i|k}$ denote the nominal state and input at prediction step $i$, computed at time $k$. This nominal trajectory may come from a reference trajectory, from the solution at the previous MPC step, or from the previous iteration of a sequential linearization procedure. Define the corresponding nominal GP query as
\[
    \bar z_{i|k}
    =
    \begin{bmatrix}
        \bar x_{i|k}^{\top} &
        \bar u_{i|k}^{\top}
    \end{bmatrix}^{\top}.
\]
The mean-only prediction model can be written as
\begin{equation}
    \mu_{i+1|k}^x
    =
    f(\mu_{i|k}^x,u_{i|k})
    +B_d\mu^d(\mu_{i|k}^x,u_{i|k}) \, .\label{eqn:mean_only_compact_model}
\end{equation}
To linearize this model, define the deviations from the nominal trajectory as
\[
    \delta x_{i|k}=\mu_{i|k}^x-\bar x_{i|k},
    \qquad
    \delta u_{i|k}=u_{i|k}-\bar u_{i|k}.
\]
Applying a first-order Taylor expansion of both the nominal model and the GP mean around $\bar z_{i|k}$ gives
\begin{align}
    \mu_{i+1|k}^x
    \approx{}&
    f(\bar x_{i|k},\bar u_{i|k})
    +\left.\frac{\partial f}{\partial x}\right|_{\bar z_{i|k}}\delta x_{i|k}
    +\left.\frac{\partial f}{\partial u}\right|_{\bar z_{i|k}}\delta u_{i|k}
    \notag\\
    &\quad
    +B_d\mu^d(\bar x_{i|k},\bar u_{i|k})
    +B_d\left.\frac{\partial \mu^d}{\partial x}\right|_{\bar z_{i|k}}\delta x_{i|k}
    +B_d\left.\frac{\partial \mu^d}{\partial u}\right|_{\bar z_{i|k}}\delta u_{i|k} \, .\label{eqn:linearized_mean_only_taylor}
\end{align}
If the nominal trajectory satisfies the mean-only model, then
\[
    \bar x_{i+1|k}
    =
    f(\bar x_{i|k},\bar u_{i|k})
    +B_d\mu^d(\bar x_{i|k},\bar u_{i|k}).
\]
Subtracting this nominal prediction from \eqref{eqn:linearized_mean_only_taylor} gives the deviation dynamics
\begin{equation}
    \delta x_{i+1|k}
    =
    A_{i|k}\delta x_{i|k}
    +B_{i|k}\delta u_{i|k},
    \quad i=0,\ldots,N-1 \, .\label{eqn:linearized_mean_only}
\end{equation}
The local Jacobian matrices are
\begin{align}
    A_{i|k}
    & =
    \left.\frac{\partial f}{\partial x}\right|_{\bar z_{i|k}}
    +B_d\left.\frac{\partial \mu^d}{\partial x}\right|_{\bar z_{i|k}},
    \notag\\
    B_{i|k}
    & =
    \left.\frac{\partial f}{\partial u}\right|_{\bar z_{i|k}}
    +B_d\left.\frac{\partial \mu^d}{\partial u}\right|_{\bar z_{i|k}} \, .\label{eqn:linearized_mean_only_jacobians}
\end{align}
Here, $A_{i|k}$ describes how a small perturbation in the predicted state affects the next predicted state, while $B_{i|k}$ describes how a small perturbation in the input affects the next predicted state. The second line of \eqref{eqn:linearized_mean_only_jacobians} is the input Jacobian and is therefore denoted by $B_{i|k}$, not by another state-transition matrix.

The step-by-step structure of the lifted model can be seen by expanding the first two prediction steps:
\begin{align*}
    \delta x_{1|k}
    & =
    A_{0|k}\delta x_{0|k}
    +B_{0|k}\delta u_{0|k},\\
    \delta x_{2|k}
    & =
    A_{1|k}A_{0|k}\delta x_{0|k}
    +A_{1|k}B_{0|k}\delta u_{0|k}
    +B_{1|k}\delta u_{1|k}.
\end{align*}
Continuing this recursion over the horizon gives a block lower-triangular lifted prediction model. Define the stacked deviation vectors
\[
    \delta\mathbf{x}_{k+1}
    =
    \begin{bmatrix}
        \delta x_{1|k}^{\top} &
        \cdots &
        \delta x_{N|k}^{\top}
    \end{bmatrix}^{\top},
    \qquad
    \delta\mathbf{u}_{k}
    =
    \begin{bmatrix}
        \delta u_{0|k}^{\top} &
        \cdots &
        \delta u_{N-1|k}^{\top}
    \end{bmatrix}^{\top}.
\]
Then the lifted state prediction can be written as
\begin{equation}
    \delta\mathbf{x}_{k+1}
    =
    \mathcal A_k\delta x_{0|k}
    +\mathcal B_k\delta\mathbf{u}_{k}.\label{eqn:lifted_state_model}
\end{equation}
Here, $\mathcal A_k$ contains products of the state-transition Jacobians from the initial deviation to each future state, while $\mathcal B_k$ contains the block lower-triangular input-to-state sensitivity matrices. This representation keeps the influence of the initial-state deviation $\delta x_{0|k}$ explicit. That term should not be dropped unless $\delta x_{0|k}=0$ by construction.

If the MPC objective is written in terms of controlled outputs, the output map can also be linearized along the nominal trajectory. Let
\[
    C_{i|k}
    =
    \left.\frac{\partial h}{\partial x}\right|_{\bar x_{i|k}},
    \qquad
    \delta y_{i|k}
    \approx
    C_{i|k}\delta x_{i|k}.
\]
Stacking the output deviations gives
\begin{equation}
    \delta\mathbf{y}_{k}
    =
    \mathcal A_{y,k}\delta x_{0|k}
    +\mathcal B_{y,k}\delta\mathbf{u}_{k},\label{eqn:lifted_output_model}
\end{equation}
where $\mathcal A_{y,k}$ and $\mathcal B_{y,k}$ are obtained by multiplying the lifted state matrices in \eqref{eqn:lifted_state_model} by the block-diagonal output Jacobian matrix formed from $C_{1|k},\ldots,C_{N|k}$.

For the unconstrained quadratic case, \eqref{eqn:lifted_output_model} converts the local GP-MPC problem into a standard least-squares update in the input deviation $\delta\mathbf{u}_k$. For example, when the input-rate penalty is omitted or absorbed into the lifted matrices, the local quadratic problem has the form
\[
    \min_{\delta\mathbf{u}_k}
    \left\|
        \tilde{\mathbf y}_{k}
        -\mathcal A_{y,k}\delta x_{0|k}
        -\mathcal B_{y,k}\delta\mathbf{u}_{k}
    \right\|_{\mathcal Q}^{2}
    +
    \left\|
        \bar{\mathbf u}_{k}
        +\delta\mathbf{u}_{k}
        -\mathbf u_{r,k}
    \right\|_{\mathcal R}^{2}.
\]
Setting the derivative with respect to $\delta\mathbf{u}_k$ to zero gives the analytical update summarized in Appendix~\ref{sec:appendix_cost_function_solve}. In a receding-horizon implementation, this update is applied iteratively to improve the nominal input sequence. Only the first optimized control input is then applied to the system, and the horizon is shifted forward.

\subsection{Practical Implementation and Complexity Notes}
\label{sec:implementation_complexity}

The formulations above are intended to provide a compact bridge from the GP mean and covariance propagation in Sec.~\ref{sec:mean_var_approx} to the application examples, rather than a complete software recipe for every plant. A typical implementation separates model preparation from receding-horizon execution. Offline, one collects transition data, evaluates the physics-based nominal model, forms residual targets, normalizes the GP inputs and residual outputs, trains either an exact or sparse GP, validates prediction accuracy and uncertainty calibration, and selects the horizon length, cost weights, constraints, and chance levels. Online, the controller receives the current state estimate and reference horizon, evaluates the GP model along the prediction horizon, solves the corresponding mean-only or constrained MPC problem, applies the first control input, and shifts the horizon forward.

\paragraph{Prototype offline--online algorithm.}
The following compact prototype summarizes the main implementation steps. It is intended as an orientation for readers who are new to GP-MPC; line-by-line, application-specific algorithms and implementation details are provided in the example papers cited in Secs.~\ref{sec:path_following} and~\ref{sec:mixed_platoon}, including GP-FBLMPC path following~\cite{wang2023learning} and constrained mixed-platoon GP-MPC and sparse GP-MPC variants~\cite{wang2024improving,wang2024learning-based,wang2024enhancing}.
\begin{enumerate}
    \item \textbf{Offline model preparation:} collect state--input--next-state transition data; evaluate the nominal model; form residual targets; normalize GP inputs and residual outputs; train an exact or sparse GP residual model; and validate both prediction accuracy and uncertainty calibration.
    \item \textbf{Offline controller design:} choose the sampling time, prediction horizon, output map, cost weights, admissible input and state sets, and, for constrained GP-MPC, chance levels and the covariance-propagation approximation.
    \item \textbf{Online initialization:} at sampling time $k$, measure or estimate the current state $\hat x_k$, construct the reference horizon $r_{i|k}$, and warm-start the input sequence from the previous MPC solution when available.
    \item \textbf{Mean-only prediction branch:} evaluate the GP posterior mean along the horizon, propagate the mean dynamics in Eq.~\eqref{eqn:mean_only_dynamics}, solve the mean-only problem in Eqs.~\eqref{eqn:mean_only_objective}--\eqref{eqn:mean_only_output}, apply the first input, shift the horizon, and repeat.
    \item \textbf{Uncertainty-aware constrained branch:} initialize $\mu_{0|k}^x$ and $\Sigma_{0|k}^x$, evaluate the GP posterior mean and covariance along the horizon, propagate covariance with the selected approximation, tighten chance constraints using Eq.~\eqref{eqn:generic_tightening}, solve the constrained problem in Eqs.~\eqref{eqn:constrained_objective}--\eqref{eqn:constrained_chance}, apply the first input, shift the horizon, and repeat.
\end{enumerate}

For the mean-only formulation in Sec.~\ref{sec:mean_only_mpc}, the online optimizer uses the GP posterior mean as a deterministic residual correction and has the structure of a deterministic nonlinear or locally quadratic MPC problem. For the constrained formulation in Sec.~\ref{sec:constrained_gp_mpc}, the online optimizer additionally propagates covariance and tightens chance constraints using expressions such as \eqref{eqn:generic_tightening}. The exact implementation details therefore depend on the selected plant model, output map, solver, and constraint set, which is why the prototype above should be read together with the application-specific algorithms in the cited implementation papers.

The main computational bottleneck is usually the repeated GP evaluation inside the MPC prediction horizon and, for nonlinear or constrained solvers, inside multiple optimization iterations. With $n$ training points, exact GP training or covariance factorization requires $\mathcal{O}(n^3)$ operations and $\mathcal{O}(n^2)$ storage, while straightforward single-query prediction requires approximately $\mathcal{O}(n)$ operations for the posterior mean and $\mathcal{O}(n^2)$ operations for the posterior variance \cite{rasmussen2006,liu2020gaussian}. Sparse inducing-point approximations with $m\ll n$ can reduce the dominant preparation cost to approximately $\mathcal{O}(nm^2)$ and the typical online mean/variance costs to approximately $\mathcal{O}(m)$ and $\mathcal{O}(m^2)$, respectively \cite{liu2020gaussian,snelson2005sparse}. Thus, a horizon of length $N$ and $n_{\mathrm{it}}$ solver iterations roughly multiplies these online GP costs by $Nn_{\mathrm{it}}$, before accounting for the solver-specific cost of the MPC optimization. This scaling explains why sparse GP models, warm starts, short horizons, linearized lifted forms, and covariance approximations are commonly used in real-time GP-MPC.

Software tools can support these steps, but they do not remove the need to match the formulation to the task. GPyTorch provides scalable GP modeling tools built on PyTorch \cite{gardner2018gpytorch}, while CasADi provides automatic differentiation and nonlinear optimization interfaces that are widely used for MPC implementation \cite{andersson2019casadi}. Recent GP-MPC survey work similarly identifies scalability, tractable prediction approximations, online adaptation, and closed-loop guarantees as central design considerations \cite{scampicchio2025gaussian}. Accordingly, this tutorial emphasizes the mathematical formulation and the interpretation of the examples, while detailed code-level choices should be made with reference to the specific application papers and software environment.

\subsection{Closed-Loop Guarantees and Scope of the Examples}
\label{sec:guarantees}

A GP residual model and its posterior covariance do not, by themselves, establish closed-loop stability, robust stability, recursive feasibility, or probabilistic constraint satisfaction. Such guarantees require assumptions on model error and uncertainty calibration together with MPC design ingredients such as invariant or terminal sets, terminal costs, backup controllers, robust/tube constructions, or explicitly verified confidence bounds. For example, GP-based MPC formulations with additional terminal and update conditions have been developed to establish recursive constraint satisfaction and stability properties \cite{maiworm2021online,koller2018safe}; recent review work also identifies scalability, tractable prediction approximations, online model updates, and theoretical guarantees as continuing challenges in GP-MPC \cite{scampicchio2025gaussian}.

Accordingly, the path-following examples in Sec.~\ref{sec:path_following} are presented as empirical demonstrations of mean-only learning-enhanced prediction and tracking performance, not as formal stability or recursive-feasibility results. The mixed-platoon example in Sec.~\ref{sec:mixed_platoon} implements explicit probabilistic distance-constraint tightening; its safety interpretation remains conditional on the Gaussian approximation, selected chance level, and fidelity of the GP uncertainty model. Table~\ref{tab:formulation_mapping} summarizes how the tutorial formulations map to the application examples.

\begin{table}[t]
\caption{Mapping from the tutorial formulations to the application examples.}
\label{tab:formulation_mapping}
\centering
\footnotesize
\begin{tabular}{p{0.24\linewidth}p{0.22\linewidth}p{0.22\linewidth}p{0.22\linewidth}}
\toprule
Application & GP information used online & Constraints in the presented implementation & Formulation classification \\
\midrule
GP-NMPC path following & Posterior mean & No explicit inequality constraints & Mean-only unconstrained GP predictive control \\
GP-FBLMPC path following & Posterior mean & No explicit inequality constraints & Mean-only unconstrained GP predictive control \\
Mixed-platoon GP-MPC & Mean and propagated variance & Spacing, speed, acceleration, and chance constraints & Uncertainty-aware constrained GP-MPC \\
Dynamic sparse GP-MPC & Sparse mean and variance & Retains the mixed-platoon constraints & Sparse constrained GP-MPC \\
\bottomrule
\end{tabular}
\end{table}

\section{Application I: Mean-Only Unconstrained GP Predictive Control for Mobile-Robot Path Following}
\label{sec:path_following}
This section relates two mobile-robot path-following studies to the mean-only unconstrained formulation in Sec.~\ref{sec:mean_only_mpc}: GP-based nonlinear MPC (GP-NMPC) \cite{ostafew2016} and GP-based feedback linearization MPC (GP-FBLMPC) \cite{wang2023learning}. In the terminology of Sec.~\ref{sec:regulation_tracking}, these are output-tracking examples: the controller regulates path-error variables or pose-related tracking errors rather than independently tracking arbitrary full-pose references. Both reported implementations use GP posterior means to improve the prediction model during path-following control; neither carries the GP posterior covariance into the online optimizer or imposes uncertainty-tightened inequality constraints. They are therefore best interpreted as empirical examples of mean-only unconstrained GP predictive control. This classification is important because the examples demonstrate learning-enhanced prediction and tracking performance, but they do not claim the chance-constraint handling or probabilistic safety interpretation introduced later in Sec.~\ref{sec:mixed_platoon}.

\subsection{GP-based Nonlinear MPC}
\label{sec:gp-nmpc}
The most direct utilization of GP models is identifying and learning the discrepancies between theoretical nominal models and actual observed behavior in robotic systems. By precisely modeling these discrepancies, GP models significantly enhance the prediction accuracy of the system's dynamics. This improvement in predictive accuracy directly contributes to superior control performance, enabling the system to adjust its actions based on a more accurate representation of its interactions with the environment.

In mobile robotics, especially for path-following tasks, employing GP models has proven to be highly effective. Path following, a fundamental functionality for autonomous vehicles and robots, requires the robot to precisely adhere to a predetermined path without temporal constraints. This challenge is particularly challenging for wheeled robots navigating complex off-road terrains, where accurately modeling the dynamics of robot-terrain interaction becomes problematic. The work presented in \cite{ostafew2016} is a pioneering application of GP-based MPC to this challenge, demonstrating a significant improvement in path-following accuracy by adapting the MPC model to reflect actual field observations through GP learning, thereby underscoring the efficacy of GP models in complex environmental interactions.

\begin{figure}
    \centering
    \includegraphics[width=0.90\columnwidth]{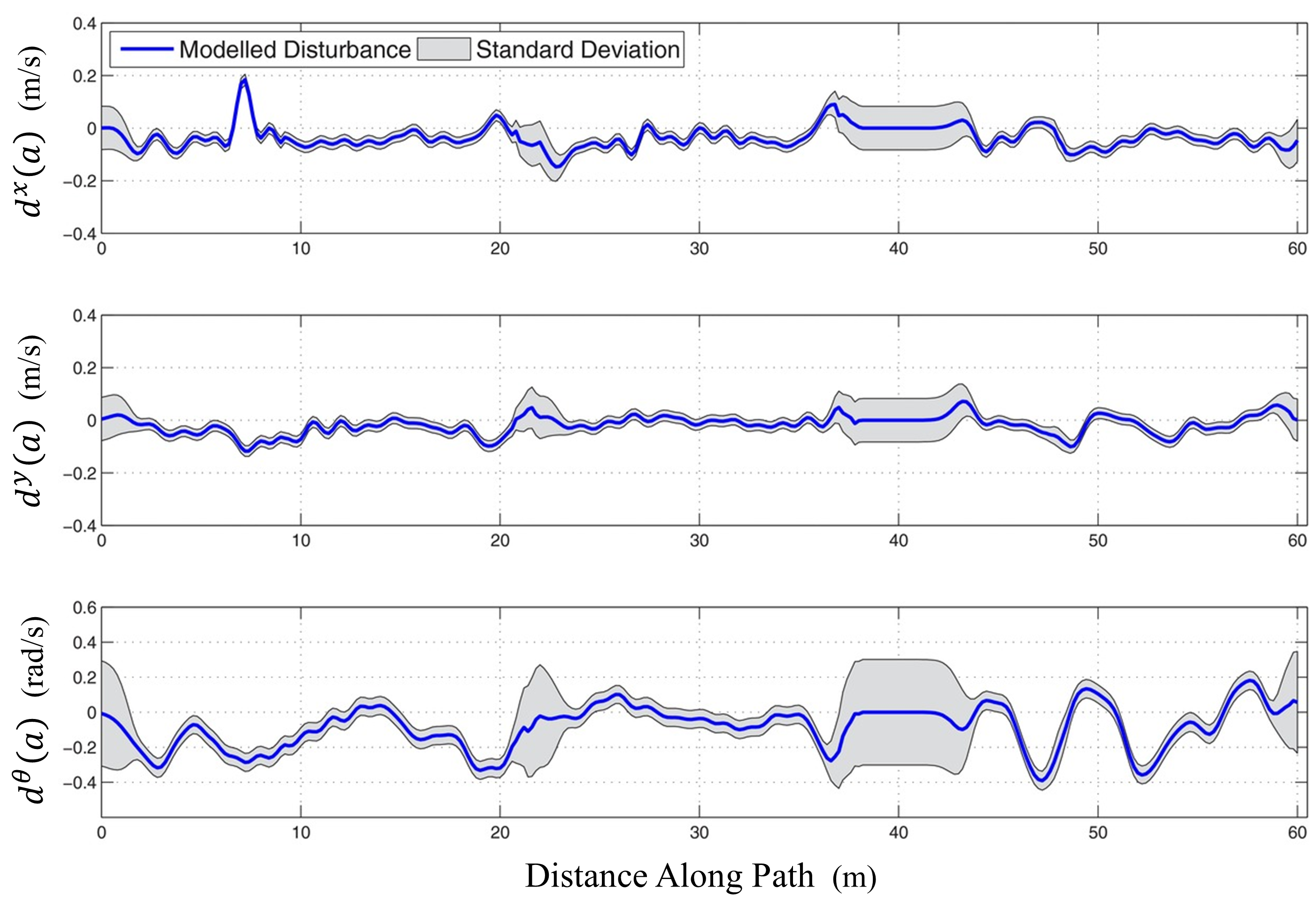}
    \caption{Illustration of the trained GP models for the robot pose variables $x$, $y$, and $\theta$ (${d}^{x}(\mathbf{a})$, ${d}^{y}(\mathbf{a})$, and ${d}^{\theta}(\mathbf{a})$ respectively), based on a 60\,m field experiment dataset \eqref{eqn:gp_dataset}. These models, developed under the assumption of uncorrelated disturbances, demonstrate the application of \eqref{eqn:gp_Bd} in creating separate GP models for each pose variable \cite{ostafew2016}.}
        \Description{A figure supporting the tutorial discussion and illustrating the corresponding GP-MPC concept or application result.}
    \label{figure:GP-NMPC_gp_modeling}
\end{figure}

A general robotic system model incorporating dynamic elements into kinematic equations can be expressed as:
\begin{equation} 
    \mathbf{x}_{k+1}=\overbrace{\mathbf{f}\left(\mathbf{x}_{k}, \mathbf{u}_{k}\right)}^{\text {nominal model }}+\overbrace{\mathbf{g}(\underbrace{\mathbf{x}_{k}, \mathbf{v}_{k-1}, \mathbf{u}_{k}, \mathbf{u}_{k-1}}_{{a}_{k}})}^{\text {GP learning model}} \, . \label{eqn:system_model_gp_nmpc}
\end{equation}
In this model, $\mathbf{x}_{k} = (x_k, y_k, \theta_k)$ denotes the robot's pose, $\mathbf{v}_{k} = (v_{k}^\text{act},\omega_{k}^\text{act})$ the actual velocity, and $\mathbf{u}_{k} = (v_{k}^\text{cmd},\omega_{k}^\text{cmd})$ the control inputs (linear and angular velocity) at time $k$. The disturbance query state is defined as:
\begin{equation}
    {a}_{k}=\left(\mathbf{x}_{k}, \mathbf{v}_{k-1}, \mathbf{u}_{k}, \mathbf{u}_{k-1}\right) \, . \label{eqn:gp_disturbance_state}
\end{equation}
Utilizing this model, at time $k-1$, discrepancies between the nominal model and actual system behavior are estimated by a GP disturbance model as:
\begin{ceqn}
    \begin{equation}
        \hat{\mathbf{g}}({a}_{k-1}) = \hat{\mathbf{x}}_{k}-\mathbf{f}\left(\hat{\mathbf{x}}_{k-1}, \mathbf{u}_{k-1}\right) \, . \label{eqn:gp_model_est}
    \end{equation}
\end{ceqn}
Here, $\hat{\mathbf{x}}_{k}$ and $\hat{\mathbf{x}}_{k-1}$ represent the robot poses from the on-board navigation system at time $k$ and $k-1$ respectively, $\mathbf{u}_{k-1}$ is the control input at time $k-1$. In the disturbance query state ${a}_{k-1}=\left(\hat{\mathbf{x}}_{k-1}, \hat{\mathbf{v}}_{k-2}, \mathbf{u}_{k-1}, \mathbf{u}_{k-2}\right)$, the $\hat{\mathbf{v}}_{k-2}$ is calculated by the estimated robot poses $\hat{\mathbf{x}}_{k-1}$ and $\hat{\mathbf{x}}_{k-2}$, which indicates the disturbance query state ${a}_{k}$ requires historic states data. In one experiment for data collection, the observed disturbance datasets are collected and prepared by following \eqref{eqn:gp_model_est} as:
\begin{ceqn}
    \begin{align}
       \mathcal{D}_m^x &= \{\mathbf{a} = [{a}_1, \cdots, {a}_{k-1}, \cdots, {a}_m]^\top, \mathbf{g}^{x}(\mathbf{a}) = [{g}^{x}({a_1}), \cdots, {g}^{x}({a_k}), \cdots, {g}^{x}({a_m})]^\top \} , \nonumber \\
       \mathcal{D}_m^y &= \{\mathbf{a} = [{a}_1, \cdots, {a}_{k-1}, \cdots, {a}_m]^\top, \mathbf{g}^{y}(\mathbf{a}) = [{g}^{y}({a_1}), \cdots, {g}^{y}({a_k}), \cdots, {g}^{y}({a_m})]^\top \} , \nonumber \\
       \mathcal{D}_m^{\theta} &= \{\mathbf{a} = [{a}_1, \cdots, {a}_{k-1}, \cdots, {a}_m]^\top, \mathbf{g}^{\theta}(\mathbf{a}) = [{g}^{\theta}({a_1}), \cdots, {g}^{\theta}({a_k}), \cdots, {g}^{\theta}({a_m})]^\top \} \, , \label{eqn:gp_dataset}
    \end{align}
\end{ceqn}
Using these datasets and \eqref{eqn:gp_Bd}, three separate GP models, ${d}^{x}(\mathbf{a})$, ${d}^{y}(\mathbf{a})$, and ${d}^{\theta}(\mathbf{a})$, are trained for each variable of the robot pose, $x$, $y$, and $\theta$ respectively, by assuming the disturbances are uncorrelated. The trained GP models with 60\,m field experiment data are shown in Fig. \ref{figure:GP-NMPC_gp_modeling} \cite{ostafew2016}. 

Utilizing the GP-enhanced prediction model \eqref{eqn:system_model_gp_nmpc}, the GP-NMPC controller in \cite{ostafew2016} evaluates the GP posterior mean within an unconstrained Gauss--Newton NMPC solve. In the notation of Sec.~\ref{sec:mean_only_mpc}, the GP term acts as a learned deterministic residual correction in the prediction model. The online solve does not propagate $\Sigma^d$ or impose the chance constraints in Sec.~\ref{sec:constrained_gp_mpc}. Through 3\,km field tests on three different robots, the study reported improved path-following accuracy in off-road environments. The nonlinear iterative optimization and the path-dependent learned pose disturbances motivate the GP-FBLMPC strategy discussed next, which reduces online computation and improves transfer to paths not represented in the training data. 

\subsection{GP-based Feedback Linearization MPC}
\label{sec:gp-fblmpc}
The GP-FBLMPC methodology \cite{wang2023learning} combines GP mean predictions with a feedback-linearized predictive-control representation. In the reported implementation, the feedback-linearized error dynamics permit a non-iterative quadratic control update, while the GP models learn residual error behavior from previous traversals. As with GP-NMPC in this section, GP-FBLMPC is a mean-only unconstrained implementation: GP covariance and explicit inequality constraints are not included in its online solve.

\begin{figure}
    \begin{center}
    \vspace{0.15cm}
        \includegraphics[trim=0cm 0cm 1.1cm 0cm, width=0.8\columnwidth]{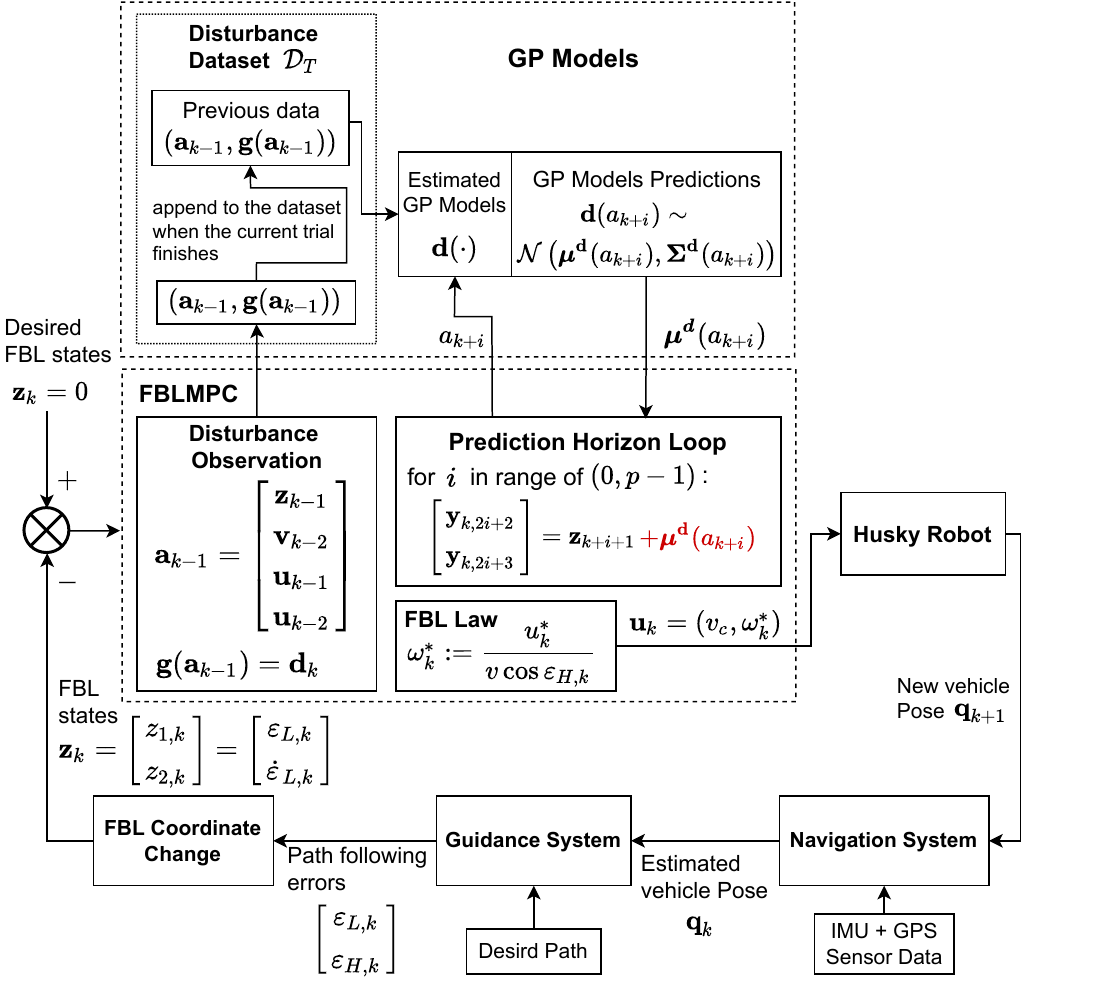}
        \caption{Block diagram of path-following control using the GP-FBLMPC algorithm. GP models capture unmodeled system dynamics from previous experiences. Estimated GP model means  ($\boldsymbol {\mu}^{\mathbf{d}}$) are integrated into the system model within the MPC prediction horizon loop \cite{wang2023learning}.}
            \Description{A figure supporting the tutorial discussion and illustrating the corresponding GP-MPC concept or application result.}
    \label{fig:gpfblmpc}
    \end{center}
\end{figure}

Figure~\ref{fig:gpfblmpc} summarizes this application-specific implementation and complements the generic workflow in Fig.~\ref{fig:gp_mpc_workflow}. The GP predictions shown in the prediction-horizon loop are posterior means integrated into a path-error controller. Thus, this example is most closely connected to the mean-only model in Sec.~\ref{sec:mean_only_mpc} and the lifted unconstrained update in Sec.~\ref{sec:mean_only_lifted}.

\subsubsection{Feedback Linearization in MPC}
Before integrating GP predictions with FBLMPC, it is useful to distinguish feedback linearization from the local first-order approximation in Eq.~\eqref{eqn:linearized_mean_only}. Feedback linearization uses a change of variables and an input transformation to obtain exact linear dynamics for selected outputs within an admissible operating region \cite{slotine1991}. For the path-error variables below, the transformation is valid only when $v\neq 0$ and $\cos\varepsilon_H\neq 0$; it is therefore not a claim of global linearization over the robot's entire operating range.

For wheeled robots with a differential drive mechanism, the discretized kinematic model is:
\begin{equation}
    \begin{bmatrix} 
            x_{k+1} \\ y_{k+1} \\ \theta_{k+1}
            \end{bmatrix} =
        \begin{bmatrix}
            {x}_{k} \\ {y}_{k} \\ {\theta}_{k}
            \end{bmatrix}
            + T
        \begin{bmatrix}
            \cos\theta_{k} & 0 \\ 
            \sin\theta_{k} & 0 \\ 
            0 & 1
        \end{bmatrix}
        \begin{bmatrix}
            v_{k} \\ 
            \omega_{k} \\ 
        \end{bmatrix} \, . \label{eqn:robot_kinematics_discret}
\end{equation} 
where $x$, $y$, and $\theta$ represent the robot's pose states, and the control inputs are the commanded forward velocity $v$ and yaw rate $\omega$. Defining the path-following errors as lateral $\varepsilon_{L} \coloneqq y$ and heading $\varepsilon_{H} \coloneqq \theta$, the instantaneous rate of change of the path-following errors are: 
\begin{ceqn}
    \begin{align} 
        \dot \varepsilon_{L} &= \dot y = v\sin\varepsilon_{H}  \, , \label{eqn:lateral_error} \\
        \dot \varepsilon_{H} &= \dot \theta = \omega \, . \label{eqn:heading_error}
    \end{align}
\end{ceqn}
The kinematic model in Eq.~\eqref{eqn:robot_kinematics_discret} is a differential-drive/unicycle-type model. The control objective is path following: the relevant outputs are lateral and heading-related path-error coordinates, rather than arbitrary independent references for all three pose states $(x,y,\theta)$. Consequently, this example does not require exact independent tracking of three pose outputs with two inputs.

The feedback linearization process begins with the introduction of new coordinates $z_{1} \equiv \varepsilon_{L} = y$ and $z_{2} \equiv \dot{\varepsilon}_{L} = \dot{y}$, designated as feedback linearization (FBL) states. This leads to a transformed system dynamics represented as:
\begin{ceqn}
    \begin{equation}
        \dot{\mathbf{z}} = 
        \begin{bmatrix}
        \dot z_{1} \\
        \dot z_{2}
        \end{bmatrix}
        =\begin{bmatrix}
        0 & 1\\
        0 & 0
        \end{bmatrix}
        \begin{bmatrix}
        z_{1} \\
        z_{2}
        \end{bmatrix}
        +
        \begin{bmatrix}
        0\\
        1
        \end{bmatrix}
        u
        = \mathbf{A}\mathbf{z} + \mathbf{B}u \, , \label{eqn:FLlinear_system}
    \end{equation}
\end{ceqn}
where $u$ is the virtual input. For constant linear velocity, $u = \dot z_{2} = v\omega \cos\varepsilon_{H}$ and $\omega = u/(v \cos\varepsilon_{H})$. The transformation requires $v \neq 0$ and $\varepsilon_H \in (-{\pi}/{2}, {\pi}/{2})$. 

The FBLMPC strategy computes a sequence of control inputs $\mathbf{u}_k$ across a finite horizon $N$ to minimize the predicted path-following errors. The control sequence and the incremental control updates are represented by $\mathbf{u}_{k} \coloneqq \Delta\mathbf{u}_{k} + \mathbf{u}_{k-1}= (u_{k}, u_{k+1},..., u_{k+N-1})$ and $\Delta \mathbf{u}_{k} = (\Delta u_{k}, \Delta u_{k+1},..., \Delta u_{k+N-1})$, respectively. By discretizing the linearized system equation \eqref{eqn:FLlinear_system}, we get:
\begin{ceqn}
    \begin{equation}
        {\mathbf{z}}_{k+1} = \mathbf{F}\mathbf{z}_k + \mathbf{G}u_k = 
            \begin{bmatrix}
                1 & T\\
                0 & 1 
            \end{bmatrix}
            \begin{bmatrix}
                z_{1,k}\\
                z_{2,k}
            \end{bmatrix}
        +   \begin{bmatrix}
                \frac{T^2}{2}\\
                T   
            \end{bmatrix} u_k \, ,  \label{eqn:discrete_lineareqn}
    \end{equation} 
\end{ceqn}
where $T>0$ is a small update interval, and $\mathbf{z}_k=\mathbf{z}(kT)$, $k=0,1,2,\ldots$. The FBL states over the prediction horizon are described by:
\begin{ceqn}
    \begin{equation} 
        \mathbf{y}_{k+1} \coloneqq \Delta \mathbf{y}_{k+1} + \mathbf{y}_{k} = (\mathbf{z}_{k+1}, \mathbf{z}_{k+2},..., \mathbf{z}_{k+N}) \, , 
        \label{eqn:y_k+1}
    \end{equation}
\end{ceqn}
where $\Delta \mathbf{y}_{k} = (\Delta \mathbf{z}_{k+1}, \Delta \mathbf{z}_{k+2},..., \Delta \mathbf{z}_{k+N})$ represents the updates of FBL states over the prediction horizon. The discrete updates in control input and FBL states are defined as $\Delta u_k \coloneqq u_{k} - u_{k-1}$ and  $\Delta \mathbf{z}_{k} \coloneqq \mathbf{z}_{k} - \mathbf{z}_{k-1}$, respectively. Utilizing \eqref{eqn:discrete_lineareqn}, the dynamics of FBL states across the prediction horizon can be expressed as:
\begin{ceqn}
    \begin{align}
        \mathbf{z}_{k+1} &= \mathbf{Fz}_k + \mathbf{G}u_{k} \, , \label{eqn:z_k+1_a} \\
        \Delta \mathbf{z}_{k+1} + \mathbf{z}_{k} &= \mathbf{F}(\Delta \mathbf{z}_{k} + \mathbf{z}_{k-1}) + \mathbf{G}(\Delta u_{k} + u_{k-1})  \, , \label{eqn:z_k+1_b} \\
        \Delta \mathbf{z}_{k+1} &= \mathbf{F}\Delta \mathbf{z}_{k} + \mathbf{G}\Delta u_{k} \, , \label{eqn:z_k+1_c} \\
        \Delta \mathbf{z}_{k+2} &= \mathbf{F}\Delta \mathbf{z}_{k+1} + \mathbf{G}\Delta u_{k+1} \, , \label{eqn:z_k+1_d} \\
        & \ \, \, \vdots \nonumber\\
        \Delta \mathbf{z}_{k+N}&= \mathbf{F}^{N}\Delta \mathbf{z}_{k} + \mathbf{F}^{N-1}\mathbf{G}\Delta u_{k} + \dots + \mathbf{G}\Delta u_{k+N-1} \, , \label{eqn:z_k+1_e}
    \end{align}
\end{ceqn}
which are rewritten in matrix form as:
\begin{ceqn}
    \begin{align}
        \underbrace{
            \begin{bmatrix}
                \Delta \mathbf{z}_{k+1}  \\
                \Delta \mathbf{z}_{k+2} \\
                \vdots\\
                \Delta \mathbf{z}_{k+N} 
            \end{bmatrix}
        }_{\Delta \mathbf{y}_{k+1}}
        = 
        \underbrace{
            \begin{bmatrix}
                \mathbf{F} \\
                \mathbf{F}^{2}\\
                \vdots\\
                \mathbf{F}^{N}
                \end{bmatrix}
        }_{\mathbf{L}}
        \Delta \mathbf{z}_{k}  + 
        \underbrace{
            \begin{bmatrix}
            \mathbf{G} & 0 & \cdots & 0 & 0\\
            \mathbf{FG} & \mathbf{G} & \cdots & 0 & 0\\
            \vdots & \vdots & \ddots & \vdots & \vdots\\
            \mathbf{F}^{N-1}\mathbf{G} & \mathbf{F}^{N-2}\mathbf{G} & \cdots & \mathbf{FG} & \mathbf{G}\\
            \end{bmatrix}
        }_{\mathbf{M}}
        \underbrace{
        \begin{bmatrix}
        \Delta u_{k}  \\
        \Delta u_{k+1} \\
        \Delta u_{k+2} \\
        \vdots\\
        \Delta u_{k+N-1} 
        \end{bmatrix}
        }_{\Delta \mathbf{u}_k} \, , \label{eqn:ctrlupdatemat}
    \end{align}
\end{ceqn}
and more succinctly as:
\begin{ceqn}
    \begin{equation}
        \Delta \mathbf{y}_{k+1}
        = \mathbf{L} \Delta \mathbf{z}_{k} 
        + \mathbf{M}\Delta \mathbf{u}_k \, . \label{eqn:ctrlupdate-simp}
    \end{equation}
\end{ceqn}

In the reported GP-FBLMPC path-following problem, the desired FBL error state is zero and the compatible virtual reference input is also zero. Its quadratic cost is therefore a special-case realization of the reference-compatible tracking objective in Eq.~\eqref{eqn:reference_cost}, enabling a consistent comparison with the reported GP-NMPC baseline:
\begin{ceqn}
    \begin{equation}
    \mathbf{J}\left(\mathbf{y}_k,\mathbf{u}_{k}\right)=\mathbf{y}_{k+1}^{\top} \mathbf{Q} \mathbf{y}_{k+1}+\mathbf{u}_{k}^{\top} \mathbf{R} \mathbf{u}_{k} \, , \label{eqn:cost-function}
    \end{equation}
\end{ceqn}
where $\mathbf{Q} \in \mathbb{R}^{2N \times 2N}$ and $\mathbf{R} \in \mathbb{R}^{N \times N}$ are the weighting matrices for path-following errors and control efforts, respectively. The term $\mathbf{y}_{k+1}$ represents the predicted FBL state $\mathbf{z}$ over the prediction horizon, and $\mathbf{u}_{k}$ is the control input sequence over the prediction horizon specified ahead of \eqref{eqn:discrete_lineareqn}. 

Integrating the equation of $\mathbf{u}_{k}$, \eqref{eqn:y_k+1}, and \eqref{eqn:ctrlupdate-simp} into the cost function \eqref{eqn:cost-function}, the optimization problem becomes:
\begin{ceqn}
    \begin{align} 
        \mathbf{J}(\mathbf{y}_k,\mathbf{u}_k,\Delta \mathbf{z}_{k},\Delta \mathbf{u}_k) =  (\mathbf{L} \Delta \mathbf{z}_{k} + \mathbf{M}\Delta \mathbf{u}_k + \mathbf{y}_{k})^{\top}\mathbf{Q}(\mathbf{L} \Delta \mathbf{z}_{k}
        + \mathbf{M}\Delta \mathbf{u}_k + \mathbf{y}_{k}) + (\Delta\mathbf{u}+\mathbf{u}_{k-1})^{\top}\mathbf{R}(\Delta\mathbf{u}+\mathbf{u}_{k-1}) \, . \label{eqn:costfuncdeltau1}
    \end{align}
\end{ceqn}
The optimization process seeks the sequence of control adjustments, $\Delta \mathbf{u}^{*}_k$, that minimizes this cost function. Given the quadratic and convex nature of $\mathbf{J}$, the optimal control sequence is derived by solving the equation ${\partial J(\Delta \mathbf{u}_k)}/{ \partial \Delta \mathbf{u}_k} = 0$ as:
\begin{equation}
    \Delta \mathbf{u}^{*}_k = -(\mathbf{M}^{\top}\mathbf{Q}\mathbf{M}+\mathbf{R})^{-1}\left(\mathbf{M}^{\top}\mathbf{Q}(\mathbf{y}_{k} + \mathbf{L}\Delta\mathbf{z}_k) + \mathbf{R}\mathbf{u}_{k-1} \right) \, .  \label{eqn:optimaldeltau}
\end{equation}

The final control sequence $\mathbf{u}^{}_k$ is then obtained by adding the optimal control adjustments to the previous control sequence: $\mathbf{u}^*_k = \Delta \mathbf{u}^{*}_k + \mathbf{u}_{k-1}$. The first element of this sequence, $u^*_k$, is applied in real-time to adjust the vehicle's steering rate, calculated as:
\begin{equation}
    \omega_k^* = \frac{u_k^*}{v \cos\varepsilon_{H,k}} \, , \label{eqn:discrete_omegacalc}
\end{equation}
where $v$ is the vehicle's linear velocity and $\varepsilon_{H,k}$ is the heading error at time $k$. Note that the matrices $\left(\mathbf{M}^{\top} \mathbf{Q} \mathbf{M} + \mathbf{R}\right)^{-1}$ and $\mathbf{M}^{\top} \mathbf{Q}$ are constants at all time steps and thus can be precomputed to significantly reduce real-time computational complexity, which is a key difference between the FBLMPC strategy and traditional nonlinear MPC. 

\subsubsection{GP-FBLMPC}
In GP-FBLMPC, the focus shifts to using feedback linearized (FBL) states $\mathbf{z}_k$ as the system variables, in contrast to the robot pose states $\mathbf{x}_k$ used in GP-NMPC. Utilizing \eqref{eqn:system_model_gp_nmpc}, the relationship between the actual and predicted robot pose state $\mathbf{x}_{k+1}$ is expressed as:
\begin{equation}
   \mathbf{x}_{k+1} - \mathbf{f}\left(\mathbf{x}_{k}, \mathbf{u}_{k}\right) = \mathbf{g}\left(\mathbf{x}_{k}, \mathbf{v}_{k-1}, \mathbf{u}_{k}, \mathbf{u}_{k-1} \right) \, . \label{eqn:gp_model_fbl}
\end{equation}
where the right side of \eqref{eqn:gp_model_fbl} are the FBL states $\mathbf{z}_{k+1}$. The GP model incorporating these FBL states is formulated as:
\begin{ceqn}
    \begin{equation}
        \mathbf{z}_{k+1}=\mathbf{g}({a}_{k}) \, , \label{eqn:gp_model_fbl_z}
    \end{equation}
\end{ceqn}
with the disturbance state defined as ${a}_{k}=\left(\mathbf{z}_{k}, \mathbf{v}_{k-1}, \mathbf{u}_{k}, \mathbf{u}_{k-1}\right)$. Like GP-NMPC, assuming uncorrelated disturbance states $z_1$ and $z_2$, two separate GP models ${d}^{z_1}(\mathbf{a})$ and ${d}^{z_2}(\mathbf{a})$, are developed for each FBL state.

The developed GP-FBLMPC method was extensively tested across three different paths in sand and grass off-road terrains. Compared with GP-NMPC, GP-FBLMPC achieved comparable path-error reduction while reducing online computation. This improvement follows from the feedback-linearized lifted representation: the main online update is a quadratic input update using the learned GP mean correction, rather than an iterative nonlinear optimization over the full pose model.

\begin{figure}
    \centering
    \vspace{0.1cm}
    \subfloat{{\includegraphics[width=0.45\columnwidth]{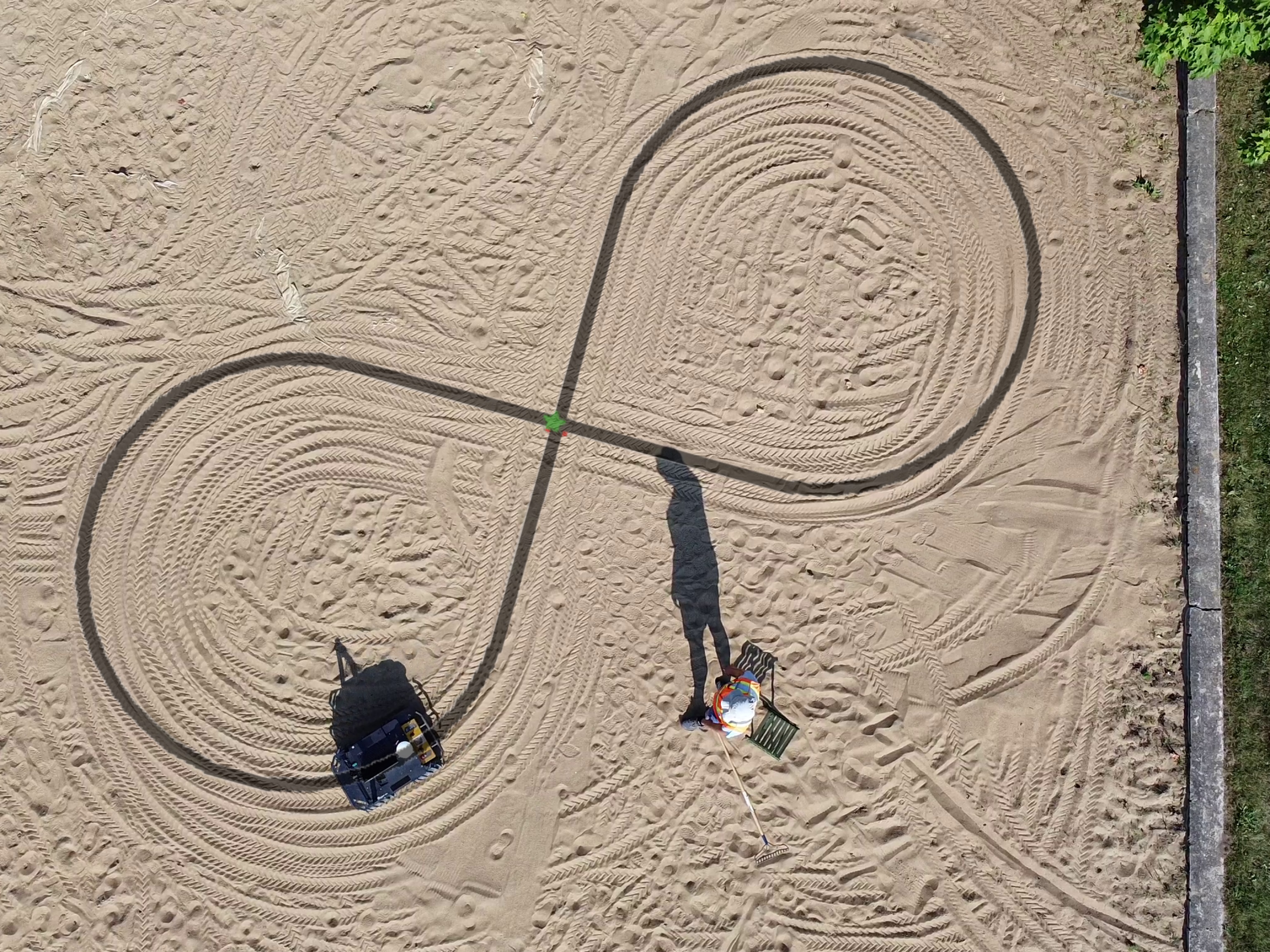} }}
    \hfill
    \subfloat{{\includegraphics[width=0.45\columnwidth]{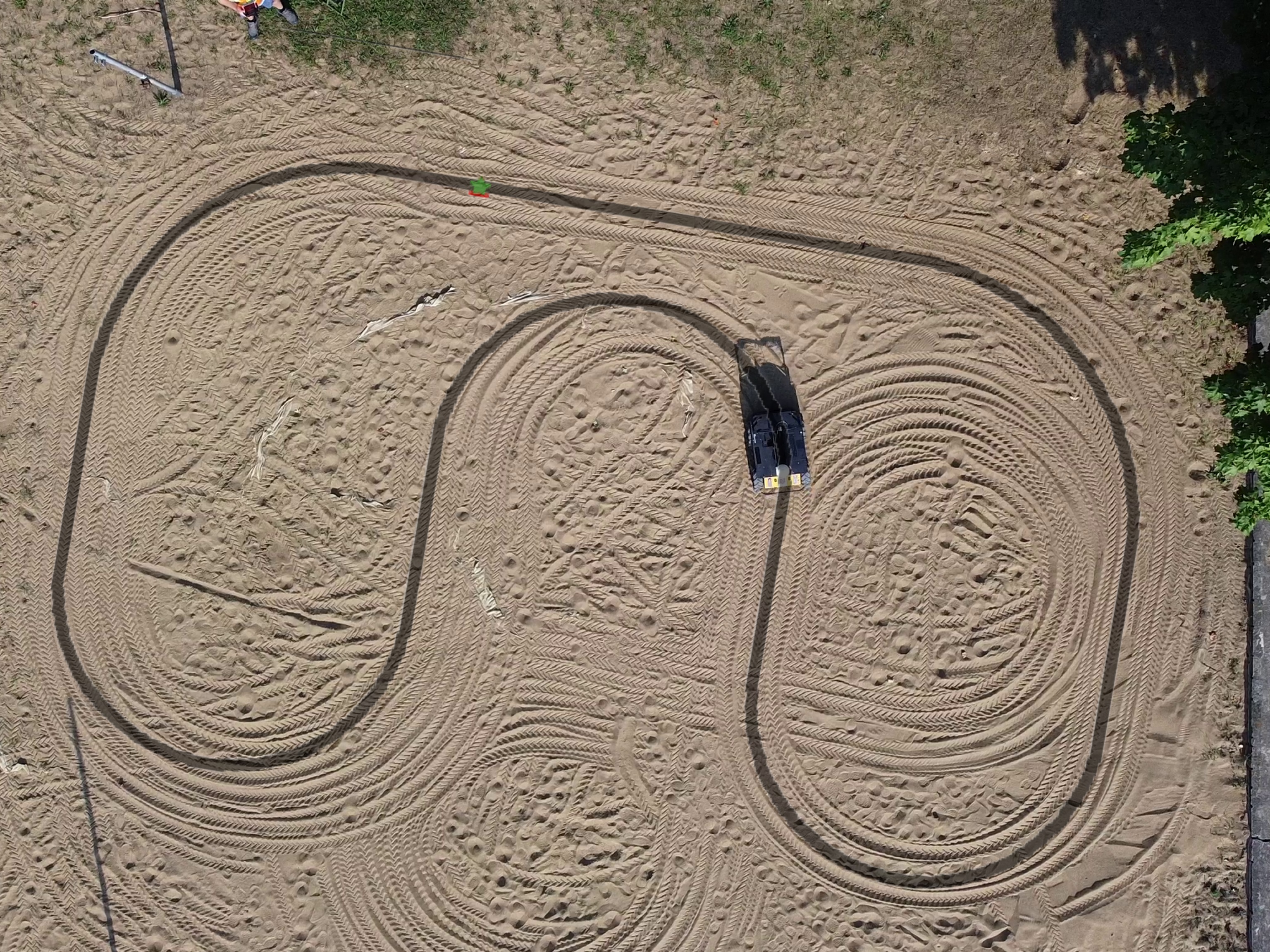} }}
    \caption{A Husky A200 robot is driven autonomously to follow the infinite path (left) and the track path (right) in sandy terrain. The desired paths, represented by black dotted lines, were blended into the field test scene image captured by a DJI Mini drone \cite{wang2023learning}.}
        \Description{Two field images of a Husky robot following an infinite path and a track path in sandy terrain.}
    \label{figure:infinite_track_path}
\end{figure}

The GP-FBLMPC method's adaptability to different paths is another key advantage, as demonstrated in a field experiment shown in Fig. \ref{figure:infinite_track_path} \cite{wang2023learning}. In this experiment, GP models, initially trained using data from an infinite path (left image in Fig. \ref{figure:infinite_track_path}), were applied to a different, track path (right image in Fig. \ref{figure:infinite_track_path}) for both GP-FBLMPC and GP-NMPC setups. This was done without re-training the GP models with data specific to the track path. The results showed that GP-FBLMPC significantly reduced path errors in comparison to basic FBLMPC and notably outperformed GP-NMPC, which demonstrated worse performance compared to the standard NMPC in this instance.

\begin{figure}
    \centering
    \vspace{0.15cm}
    \hspace{-0.1cm}\subfloat{\includegraphics[trim=1.2cm 0.4cm 1.6cm 1.1cm, width=0.75\columnwidth]{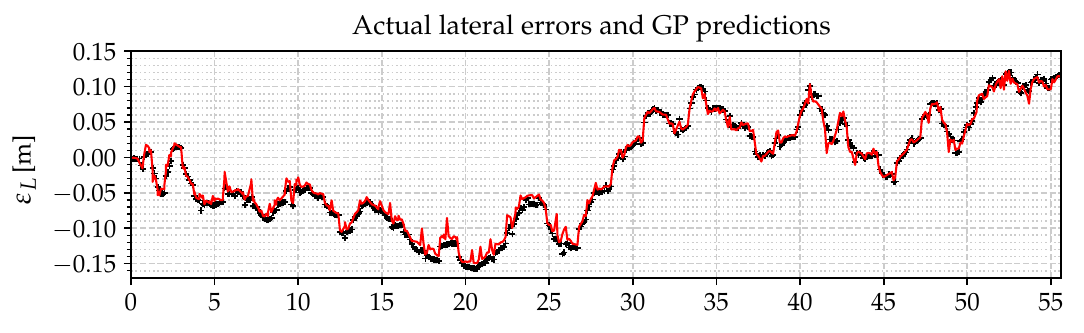}}
    \vspace{0.7cm}
    \hspace{0.0cm}\subfloat{\includegraphics[trim=1.2cm 0.4cm 1.6cm 1.1cm, width=0.74\columnwidth]{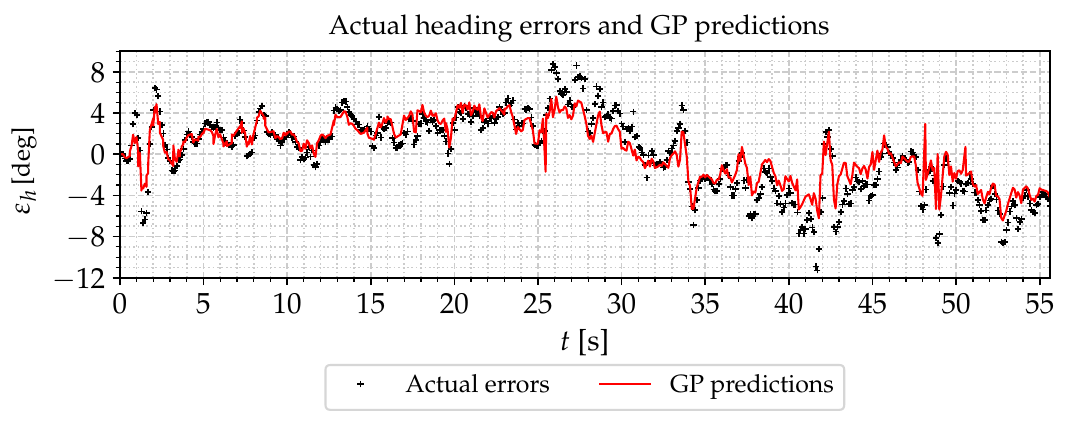}}
    \caption{Graphs depicting the lateral and heading error predictions from the GP models and the actual observed errors over time, for the GP-FBLMPC applied on the track path using GP models trained on the infinite path \cite{wang2023learning}.
    }
        \Description{Time-series plots comparing GP-predicted and observed lateral and heading path-following errors.}
    \label{figure:gp_est_track_fbl_el_eh_E4}
\end{figure}

Fig.~\ref{figure:gp_est_track_fbl_el_eh_E4} illustrates the alignment of GP predictions from GP-FBLMPC with observed lateral and heading errors on the track path. In this experiment, predictions remained accurate despite training on a different path, providing empirical evidence of path-transfer capability for the selected error-state representation; it does not constitute a formal robustness guarantee.

\section{Application II: Constrained Uncertainty-Aware GP-MPC for Mixed-Vehicle Platoons}
\label{sec:mixed_platoon}
This section presents the constrained implementation that complements the mean-only path-following examples in Sec.~\ref{sec:path_following}. The controller uses a GP-based model of human-driven vehicle (HV) behavior, propagates the predicted HV-position uncertainty, and tightens a probabilistic safe-distance constraint while also imposing deterministic speed and acceleration bounds. It is therefore an application of the uncertainty-aware constrained GP-MPC formulation in Sec.~\ref{sec:constrained_gp_mpc}. The main purpose of this section is to show where the propagated covariance enters the MPC constraints, not to claim that GP uncertainty alone provides a formal closed-loop safety guarantee.

\subsection{Chance-Constrained GP-based MPC}
\label{sec:gp_mixed_platoon}
The integration of autonomous vehicles (AVs) into existing traffic systems, especially in the domain of AV platooning, holds transformative potential for public traffic management through synchronized vehicle movements. However, the coexistence of human-driven vehicles (HVs) alongside AVs in these mixed-traffic environments introduces complex challenges. Notably, the interaction between AVs and HVs has led to an increased rate of accidents that are predominantly caused by HVs rear-ending AVs, which underscores the urgent need for innovative control strategies tailored to mixed-traffic scenarios.

\begin{figure}
    \centerline{\includegraphics[width=0.65\columnwidth]{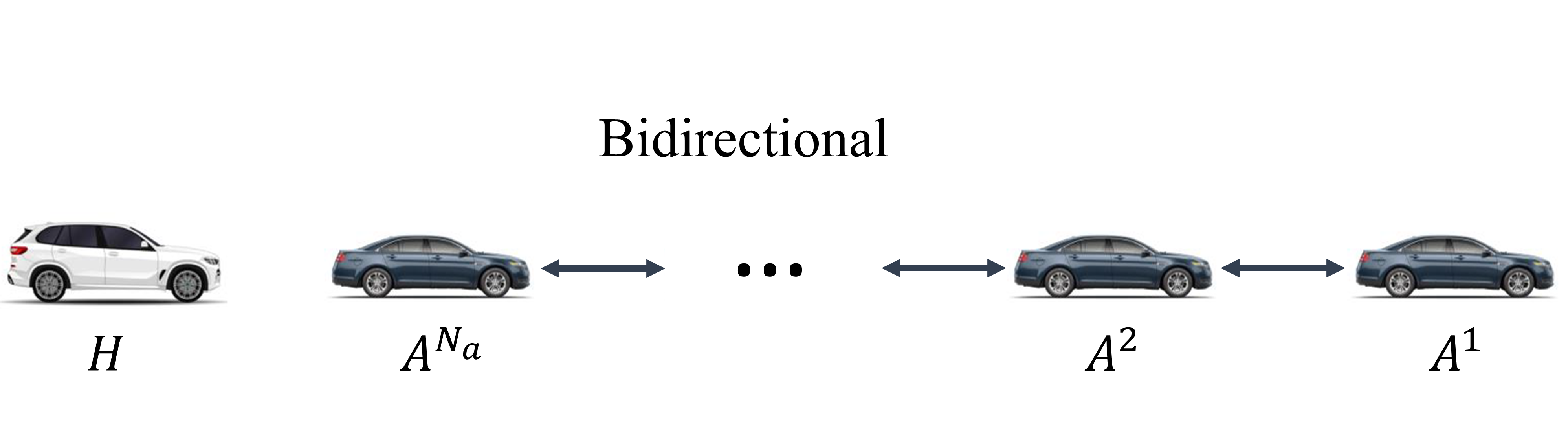}}
    \caption{A mixed vehicle platoon consists of $N_a$ connected AVs, labeled as ${A^1, A^2, \cdots, A^{N_a}}$, followed by a HV $H$. These AVs use a sequential bidirectional communication topology for data exchange, excluding direct communication with the HV. This configuration is motivated by recent studies highlighting that most accidents in mixed traffic involve HVs rear-ending AVs, leading to the specific platoon arrangement showcased \cite{wang2024improving}.
    }
        \Description{A mixed-vehicle platoon diagram with connected autonomous vehicles followed by a human-driven vehicle.}
    \label{figure:mixed_platoon}
\end{figure}

Addressing this gap, \cite{wang2024improving} proposed a GP learning-based MPC strategy for longitudinal car-following control within mixed-vehicle platoons, as illustrated in Fig.~\ref{figure:mixed_platoon}. The method incorporates predicted HV uncertainty so that the required safe distance can be enlarged as uncertainty increases. The referenced study compares this uncertainty-aware constrained controller with standard MPC in scenarios including emergency braking; the present tutorial focuses on how its GP uncertainty is integrated into the MPC constraints.

\subsubsection{HV Modeling}
In longitudinal car-following scenarios, HVs are traditionally modeled using a nominal function with a fixed human reaction time delay to emulate general human driving behaviors. A common model used for this purpose is an autoregressive with exogenous input (ARX) model, as described in \cite{wang2024improving}. The ARX model for HV velocity prediction is represented as:
\begin{ceqn} 
    \begin{align}
        v^{H}_{k} &= -c_1 v^{H}_{k-1} - c_2 v^{H}_{k-2} - c_3 v^{H}_{k-3} - c_4 v^{H}_{k-4} + b_1 v^{N_a}_{k-1} + b_2 v^{N_a}_{k-2} + b_3 v^{N_a}_{k-3} + b_4 v^{N_a}_{k-4} \, , \\
        & = {f}\left(v_{k-1:k-4}^{H}, {v}_{k-1:k-4}^{N_a} \right) \, . \label{eqn:arx}
    \end{align}
\end{ceqn}
Here, $v^{H}_{k-i}$ and $v^{N_a}_{k-i}$ represent the velocities of the HV and the last vehicle in the AV platoon at time step $k-i$, respectively. The coefficients $c_{1,\cdots, 4}$ and $b_{1,\cdots, 4}$ are constant coefficients associated with velocities of the HV and the last vehicle in the AV platoon at previous time steps respectively. Similar to combining GP models with the nominal model in these path-following control application examples, GP models are combined with the nominal model \eqref{eqn:arx} not only to enhance the accuracy of the HV model but also to offer a quantifiable measure of modeling uncertainties. The GP-enhanced ARX model is developed as: 
\begin{ceqn}
    \begin{align}
        v^H_k &= \sum_{i=1}^4 -c_i v^H_{k-i} + \sum_{i=1}^4 b_i v^{N_a}_{k-i}  = {f}(\cdot) \, , \label{eqn:arxgp_a}\\
        \tilde{v}^{H}_{k}&=\overbrace{v^H_k}^{\text {ARX prediction}}+\overbrace{{g}(v^{H}_{k-1}, v^{N_a}_{k-1})} ^{\text {GP-based correction}} \, . \label{eqn:arxgp_b}
    \end{align}
\end{ceqn}
In this model, $\tilde{v}^{H}_{k}$ represents the GP-compensated velocity prediction of the HV. The GP model ${g}(\cdot)$ learns the divergence between the actual system behaviors and the ARX predictions. It considers both $v^{H}$ (HV velocity) and $v^{N_a}$ (velocity of the last AV). This ensures that the GP model adequately captures the dynamics of the HV as influenced by the actions of the leading AV.

Data for developing the HV model were collected in a human-in-the-loop Unity driving simulator, where drivers were asked to follow an AV platoon while answering algebraic questions to mimic distracted-driving conditions.
Comparative analysis using root mean square error (RMSE) against actual HV velocity data showed that the GP-enhanced HV model \eqref{eqn:arxgp_b} yielded approximately a 35.64\% improvement in prediction accuracy over the standard ARX model, highlighting its effectiveness in modeling HV behavior in mixed-vehicle platoons.

\subsubsection{Mixed Platooning Model}
To effectively manage longitudinal tracking in mixed-vehicle fleets, a GP learning-based MPC strategy is developed, utilizing the GP-enhanced HV model \eqref{eqn:arxgp_b}. This strategy integrates crucial constraints like acceleration, speed, and safe distance. A key aspect of this approach involves maintaining a safe distance within the mixed platoon. For AVs in the platoon, defined by the set $\mathbf{n_a} = {1, 2, \cdots, N_a}$, the distance between consecutive vehicles must satisfy a minimum threshold $\Delta$, i.e., $p_{k}^{\mathbf{n_a}-1}-p_{k}^{\mathbf{n_a}} \geq \Delta$. Here, $N_a$ is the number of AVs, and $p_{k}$ indicates their position. The kinematics of AVs is formulated as:
\begin{ceqn}
    \begin{align}  
    v_{k+1}^{\mathbf{n_a}} &= v_{k}^{\mathbf{n_a}} + T \, \mathrm{acc}_{k}^{\mathbf{n_a}} \, , \label{eqn:av_a} \\
    p_{k+1}^{\mathbf{n_a}} &= p_{k}^{\mathbf{n_a}} + T \, v_{k}^{\mathbf{n_a}}  \, . \label{eqn:av_b}
    \end{align}
\end{ceqn}
where $0<T \ll 1$ indicates the sample time, and $v_{k}^{\mathbf{n_a}}$ and $\mathrm{acc}_{k}^{\mathbf{n_a}}$ denote the velocity and acceleration of $A^{\mathbf{n_a}}$ (shown in Fig. \ref{figure:mixed_platoon}) respectively. AVs are assumed to be deterministic, with their states measured and communicated error-free, i.e., $\Sigma(v_{k}^\mathbf{n_a}) = 0$. By applying \eqref{eqn:arxgp_b}, the HV model is derived as follows:
\begin{ceqn} 
    \begin{align} 
        \tilde{v}_{k}^{H} &= v_{k}^{H} + {d}(v_{k-1}^{H}, {v}_{k-1}^{N_a}) \, , \label{eqn:sys_model_a} \\
        p_{k+1}^{H} &= p_{k}^{H} + T \, \tilde{v}_{k}^{H} \, , \nonumber \\ 
        &= p_{k}^{H} + T \, v_{k}^{H} + T \, {d}(v_{k-1}^{H}, {v}_{k-1}^{N_a}) \, . \label{eqn:sys_model_b}
    \end{align}
\end{ceqn}
In this model, $\tilde{v}_{k}^{H}$ is the GP-enhanced HV velocity prediction, where $v_{k}^{H}$ is the nominal velocity (computed using \eqref{eqn:arx}), and $p_{k}^{H}$ is the position state. Using \eqref{eqn:sys_model_b} and \eqref{eqn:mean_prop_final_again}, the propagation of the HV position mean can then be derived as:
\begin{equation}
    \mu \left(p_{k+1}^{H} \right) = \mu \left( p_{k}^{H} \right) + T \, \mu \left( v_{k}^{H} \right) + T \, \mu^d \left( v_{k-1}^{H}, {v}_{k-1}^{N_a} \right) \, . \label{eqn:mixed_vel_prop_c} 
\end{equation} 
Denoting $\mu \left(p_{k+1}^{H} \right)$ as $\mu_{k+1}^{p^{H}}$ and rewriting other terms similarly, \eqref{eqn:mixed_vel_prop_c} is expressed more compactly as:
\begin{ceqn} 
    \begin{align} 
        \mu_{k+1}^{p^{H}} &= \mu_{k}^{p^{H}} + T \, v_{k}^{H} + T \, \mu^d \left( v_{k-1}^{H}, {v}_{k-1}^{N_a} \right) \, , \label{eqn:mixed_vel_prop_d}
    \end{align}
\end{ceqn}
with the initial value $\mu_{0}^{p^{H}} = p_{k}^{H}$. Using \eqref{eqn:sys_model_b} and \eqref{eqn:linear_cov_mean_equiv}, the propagation of the HV position variance is derived as:
\begin{equation}
    \Sigma \left(p_{k+1}^{H} \right) = \Sigma (p_{k}^{H}) + T^2 \Sigma^d \left(v_{k-1}^{H}, {v}_{k-1}^{N_a}\right) \, . \label{eqn:var_prop_pH} 
\end{equation} 
Similarly, this can be compactly written as:
\begin{ceqn} 
    \begin{align} 
        \Sigma_{k+1}^{p^{H}} = \Sigma_{k}^{p^{H}} + T^2 \Sigma^d \left(v_{k-1}^{H}, {v}_{k-1}^{N_a}\right) \, . \label{eqn:mixed_HV_var}
    \end{align}
\end{ceqn}
with $\Sigma_{0}^{p^{H}} = 0$, this equation propagates the HV position variance used in the safety-distance tightening. This implementation adopts the mean-equivalent approximation in Eq.~\eqref{eqn:var_prop_mean_equiv}: the AV states are treated as deterministic, and cross-covariances between HV position, nominal ARX velocity, and the GP residual are neglected. Stating this approximation makes explicit how the constrained application maps to the generic propagation choices summarized in Sec.~\ref{sec:mean_var_summary} and to the constrained formulation in Sec.~\ref{sec:constrained_gp_mpc}.

\subsubsection{Safe Distance Probability Constraint}
To ensure a safe distance within a mixed-vehicle platoon, a probabilistic constraint, also known as a chance constraint, is established. This constraint is crucial for maintaining a safe separation between the trailing AV and the HV, particularly given the uncertain nature of HV behavior. The constraint is formulated as:
\begin{equation} 
    \mathrm{Pr}\left(p_{k}^{N_a}-(p_{k}^{H} + \Delta) > \Delta_\text{ext} \right) \geq p_{\text{def}} \, . \label{eqn:chance_constraint} 
\end{equation}
In this formulation, $\Delta$ is the predefined safe distance between the AVs, and $\Delta_\text{ext} \geq 0$ is an additional buffer to accommodate the stochastic behavior of the HV. The predefined $p_{\text{def}}$ denotes the desired satisfaction probability. 

To express this probabilistic constraint in a more deterministic form, we use the concept of a half-space constraint $\mathcal{X}^{hs} := \bigl\{x \vert h^{\top}x \leq b \bigl\}$, where $h \in \mathbb{R}^n$ and $b \in \mathbb{R}$. The tightened constraint on the state mean, as derived in \cite{hewing2019}, is expressed as:
\begin{equation} 
    \mathcal{X}^{h s}\left(\Sigma_{i}^{x}\right):=\left\{x \mid h^{\top} x \leq b-\Phi^{-1}\left(p_{\text{def}}\right) \sqrt{h^{\top} \Sigma_{i}^{x} h}\right\} \, . \label{eqn:half_space_constraint} 
\end{equation}
In this equation, $h^{\top} = \begin{bmatrix} -1 & 1\end{bmatrix}$, $x := \begin{bmatrix} p_{k}^{N_a} & p_{k}^{H}+\Delta \end{bmatrix}^{\top}$, and $b = -\Delta_\text{ext}$. Here, $\Phi^{-1}$ is the inverse cumulative distribution function of the standard normal distribution. Because $\Delta$ and $\Delta_\text{ext}$ are deterministic distances, they do not enter the covariance. Under the stated deterministic-AV assumption,
\begin{equation}
    \Sigma_{k}^{p}=\operatorname{cov}\!\left(\begin{bmatrix}p_k^{N_a}\\p_k^{H}\end{bmatrix}\right)
    =\begin{bmatrix}0 & 0\\0 & \Sigma_{k}^{p^{H}}\end{bmatrix} \, . \label{eqn:phi_cdf}
\end{equation}
Using the HV position variance from Eq.~\eqref{eqn:mixed_HV_var}, insertion of Eq.~\eqref{eqn:phi_cdf} into Eq.~\eqref{eqn:half_space_constraint} yields the tightened deterministic separation constraint
\begin{equation} 
    p_{k}^{N_a}-p_{k}^{H} \geq \Delta + \Delta_\text{ext} + \Phi^{-1}\left(p_{\text{def}}\right) \sqrt{ \Sigma_{k}^{p^{H}}} \, . \label{eqn:safe_dis_HV} 
\end{equation}

The chance constraint in Eq.~\eqref{eqn:chance_constraint} is thus represented deterministically by Eq.~\eqref{eqn:safe_dis_HV}. The uncertainty-dependent term increases the required separation as the predicted HV-position variance grows. Its probabilistic interpretation is conditional on the Gaussian approximation, the chosen probability level $p_\text{def}$, and the calibration of the GP uncertainty model; it is not, by itself, a deterministic safety guarantee.

\subsubsection{GP-Based MPC}
\label{sec:GP-MPC}
The resulting GP-based MPC strategy is tailored to the mixed-vehicle platoon in Fig.~\ref{figure:mixed_platoon}, where $N_a$ AVs are followed by an HV. The optimization variables include AV velocities, AV positions, AV accelerations, the predicted HV velocity, and the propagated mean and variance of the HV position. In contrast to the mobile-robot examples in Sec.~\ref{sec:path_following}, this formulation contains explicit inequality constraints. The strategy can be expressed as:
\begin{ceqn}
    \begin{align} 
       \underset{\mathbb{V}}{\text{min}} \sum_{\mathbf{n_a}=1}^{N_a} \sum_{i=k}^{k+N-1} \Big\| \mathrm{acc}_{{i}|k}^{\mathbf{n_a}} \Big\|^2_R &+ \sum_{i=k}^{k+N} \Big\| v _{{i+1}|k}^1 - v_{{i+1}|k}^\text{ref} \Big\|^2_{Q_1} + \sum_{\mathbf{n_a}=2}^{N_a}\sum_{i=k}^{k+N} \Big\| v_{{i+1}|k}^{\mathbf{n_a}} - v_{{i+1}|k}^{\mathbf{n_a}-1} \Big\|^2_{Q_2} \label{eqn:mpc_a} \\
       \text{with} \ \mathbb{V} = & \left\{v_{{i}|k}^1, v_{{i}|k}^{\mathbf{n_a}}, v_{{i}|k}^{H}, p_{{i}|k}^{\mathbf{n_a}}, \mu_{{i}|k}^{p^{H}}, \Sigma_{{i}|k}^{p^{H}}, \mathrm{acc}_{{i}|k}^{\mathbf{n_a}} \right\} \, \nonumber \\
        \text {subject to} \nonumber \\
        v_{{i+1}|k}^{\mathbf{n_a}} &= v_{{i}|k}^{\mathbf{n_a}} + T \, \mathrm{acc}_{{i}|k}^{\mathbf{n_a}} , \  p_{{i+1}|k}^{\mathbf{n_a}} = p_{{i}|k}^{\mathbf{n_a}} + T \, v_{{i}|k}^{\mathbf{n_a}} \, , \nonumber \\
        \mathbf{n_a} &= \{1, 2, \cdots, N_a\}  \, , \label{eqn:mpc_b} \\ 
        v^{H}_{{i}|k} &= {f}\left(v_{i-1:i-4 | k}^{H}, {v}_{i-1:i-4 | k}^{N_a} \right) \, , \label{eqn:mpc_c}\\
        \mu_{{i+1}|k}^{p^{H}} &= \mu_{{i}|k}^{p^{H}} + T \, v_{{i}|k}^{H} + T \, \mu^d (v_{{i-1}|{k}}^{H}, {v}_{{i-1}|{k}}^{N_a}) \, , \label{eqn:mpc_d}\\
        \Sigma_{{i+1}|k}^{p^{H}} &= \Sigma_{{i}|k}^{p^{H}} + T^2 \Sigma^d (v_{{i-1}|{k}}^{H}, {v}_{{i-1}|{k}}^{N_a}) \, , \label{eqn:mpc_e}\\
        p_{{i}|k}^{\mathbf{n_a}-1} & - p_{{i}|k}^{\mathbf{n_a}} \geq \Delta \, , \label{eqn:mpc_f}\\
        p_{{i}|k}^{N_a} & - \mu_{{i}|k}^{p^{H}} \geq \Delta + \Delta_\text{ext} + \Phi^{-1}\left(p_{\text{def}}\right) \sqrt{ \Sigma_{{i}|k}^{p^{H}}} \, , \label{eqn:mpc_g}\\
        v_{\text{min}} & \leq v_{{i}|k}^{\mathbf{n_a}} \leq v_{\text{max}} \, , \ \mathrm{acc}_{\text{min}} \leq \mathrm{acc}_{{i}|k}^{\mathbf{n_a}} \leq \mathrm{acc}_{\text{max}} \, . \label{eqn:mpc_h}
    \end{align}
\end{ceqn}
Equation~\eqref{eqn:mpc_a} minimizes AV acceleration, reference-speed tracking error, and velocity disagreement across AVs. Equations~\eqref{eqn:mpc_b}--\eqref{eqn:mpc_e} propagate deterministic AV states together with the mean-equivalent GP-based HV prediction. Equation~\eqref{eqn:mpc_g} is the uncertainty-tightened HV--AV separation constraint derived from Eq.~\eqref{eqn:safe_dis_HV}, and Eq.~\eqref{eqn:mpc_h} imposes explicit speed and acceleration limits. Consequently, Eqs.~\eqref{eqn:mpc_a}--\eqref{eqn:mpc_h} provide a concrete constrained GP-MPC instance of the generic formulation in Eqs.~\eqref{eqn:constrained_objective}--\eqref{eqn:generic_tightening}: the GP posterior mean enters the HV prediction model, and the propagated GP variance enters the tightened distance constraint.

In the tests reported in \cite{wang2024improving}, the constrained GP-MPC was benchmarked against a standard MPC in constant-velocity tracking and emergency-braking scenarios. The study reported increased safety margins and higher average speeds when the GP-based HV uncertainty was included in the spacing constraint. These results are empirical comparisons; the interpretation of closed-loop guarantees remains subject to the qualifications in Sec.~\ref{sec:guarantees}.

\subsection{Sparse Constrained GP-based MPC}
\label{sec:sparse_gp_mpc}
The constrained formulation above shows how GP uncertainty can be used inside an MPC constraint, but it also highlights the computational burden of uncertainty-aware GP-MPC. As discussed in Sec.~\ref{sec:implementation_complexity}, repeated GP mean and variance evaluations over the horizon can dominate the online cost, especially when the training dataset is large. In the mixed-platoon implementation based on a standard GP model, the reported computation time reached approximately 19 seconds per time step, which is not suitable for real-time control.

Sparse GP models provide one way to retain the constrained uncertainty-aware structure while reducing the cost of GP prediction. This distinction is important for the tutorial: the sparse variant is not a return to a mean-only approximation. Instead, it keeps the mixed-platoon constraints and still propagates a GP variance for distance tightening, but uses a sparse GP representation to reduce the cost of computing the mean and variance terms in Eqs.~\eqref{eqn:mpc_d} and \eqref{eqn:mpc_e}. Thus, the sparse controller remains within the constrained GP-MPC family summarized in Table~\ref{tab:formulation_mapping}.

\subsubsection{Sparse GP+ARX Modeling} 
The research detailed in \cite{wang2024learning-based} illustrates a successful application of the fully independent conditional (FIC) approximation, a leading-edge sparse GP approach, for enhancing computational efficiency in dynamic systems modeling. The hyperparameters for the sparse GP model were directly loaded from the trained standard GP model, and 20 inducing points were automatically selected by the FIC method within the training datasets. This incorporation resulted in a remarkable boost in computational speed, where the sparse GP model required only about 5.6\% of the average prediction time needed by the standard GP model.

\begin{figure}
    \centering
    \vspace{0.1cm}
    \includegraphics[trim=0cm 0cm 0cm 0cm, width=0.88\columnwidth]{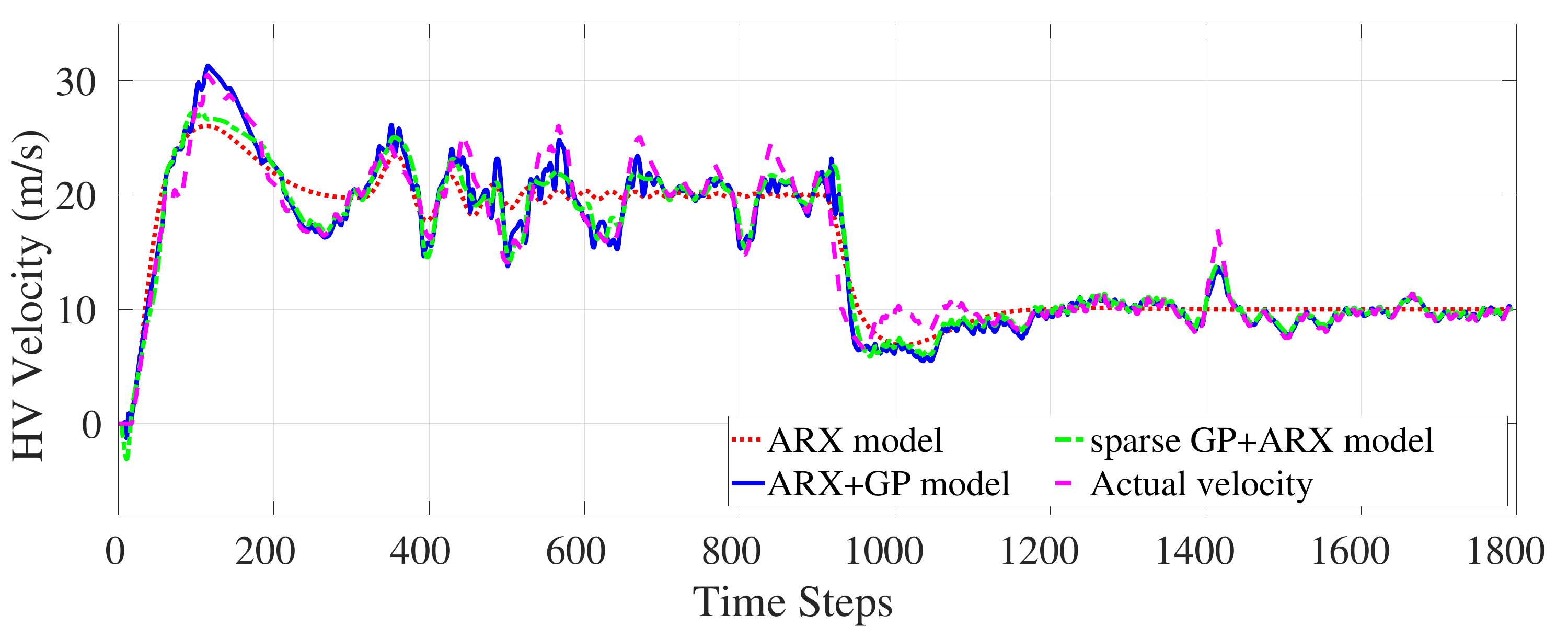}
    \caption{Comparative analysis of velocity prediction models. Plots show velocity predictions of the ARX, ARX+GP, and sparse GP+ARX models, against actual HV velocity measurements. Compared to the ARX model, the ARX+GP exhibits approximately 35.64\% improvement, and sparse GP+ARX achieves a 23.94\% enhancement with significantly reduced computational demands \cite{wang2024learning-based}. }
        \Description{Velocity prediction comparison among ARX, ARX with GP correction, sparse GP with ARX, and measured human-vehicle velocity.}
    \label{figure:gp_sparse}
\end{figure}

In terms of modeling precision, the standard GP+ARX model exhibited a notable 35.64\% improvement in forecasting HV velocities relative to the ARX nominal model. The sparse GP+ARX model achieved an approximate 23.94\% improvement compared to the ARX model, which provides a good balance of prediction accuracy and computational efficiency. To illustrate this improvement, Fig. \ref{figure:gp_sparse} displays a comparison of velocity predictions using the ARX model, the ARX+GP model, and the sparse GP+ARX model against measured HV velocities.
The comparison clearly demonstrates that both the standard GP+ARX and the sparse GP+ARX models significantly outperform the ARX model in accurately predicting the HV behaviors. These results highlight the sparse GP+ARX model's effectiveness in providing a good balance between enhanced modeling precision and substantial gains in computational efficiency.

\subsubsection{Dynamic Sparse GP Prediction in MPC}
\label{sec:dynamic_sparseGP_mpc}
Despite the notable reduction in computation time achieved by employing the sparse GP+ARX model for HV modeling, real-time integration of GP models within the MPC framework continues to pose challenges due to the inherently complex nature of MPC \cite{abdolhosseini2013efficient}. MPC involves solving an optimal control problem at each time step, adhering to constraints as outlined in equations \eqref{eqn:mpc_b}--\eqref{eqn:mpc_h}. Addressing this challenge, \cite{wang2024learning-based} introduced a dynamic sparse approximation method for GP-MPC, aimed at further speeding up the process.

Leveraging the receding-horizon characteristic of MPC, where the current predicted trajectory is initialized from the previously optimized trajectory, the dynamic sparse GP method uses the prior trajectory to update the sparse GP mean and variance required in Eqs.~\eqref{eqn:mpc_d} and \eqref{eqn:mpc_e}. This reduces repeated GP predictions across the horizon while retaining the constrained optimization conditions in Eqs.~\eqref{eqn:mpc_f}--\eqref{eqn:mpc_h}. The example is therefore classified in Table~\ref{tab:formulation_mapping} as sparse constrained GP-MPC, rather than as a mean-only approximation.

\subsubsection{Testing with Realistic Velocity Profiles}
\begin{figure}
    \vspace{0.15cm}
    \centerline{\includegraphics[width=0.85\columnwidth]{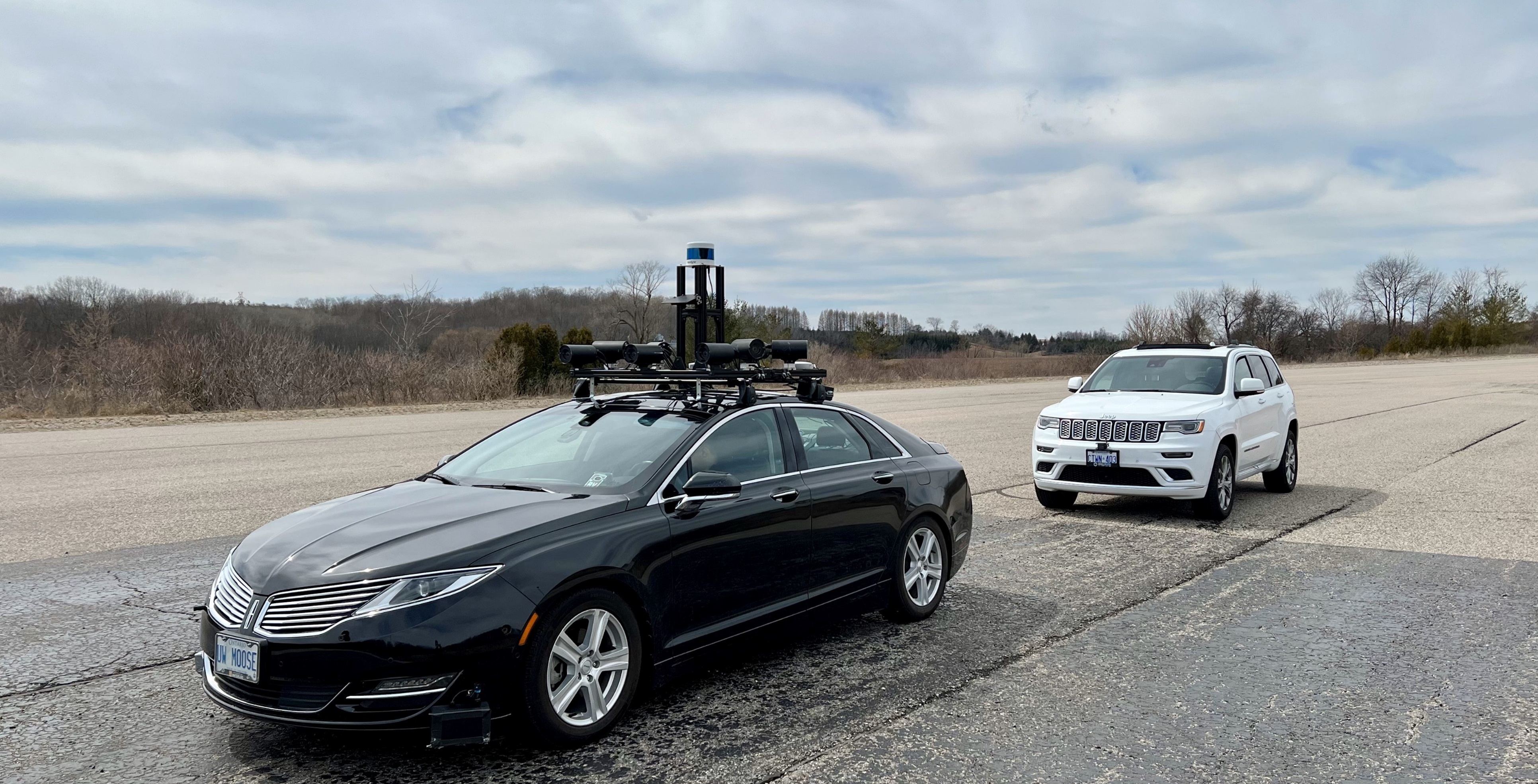}}
    \caption{The autonomous vehicle platform, UW Moose, and a following human-driven vehicle on the test track \cite{wang2024enhancing}.
    }
        \Description{An autonomous vehicle platform and a following human-driven vehicle on a test track.}
    \label{figure:AV_HV_cars}
\end{figure}

In \cite{wang2024improving} and \cite{wang2024learning-based}, the HV model including both the nominal ARX and GP components, was estimated with data collected in a human-in-the-loop Unity driving simulator. In \cite{wang2024enhancing}, the HV model incorporates field experiment data, complementing the nominal model estimated with data from the simulator. This advancement involved collecting data for the GP component during field experiments where an HV followed an AV, as shown in Fig. \ref{figure:AV_HV_cars}. The data was gathered on an 850-meter real-world track, distinguished by its varying curvatures and bumps, to provide a detailed and realistic dataset. This dataset was pivotal in improving the GP model's accuracy and reliability. Notably, compared to the simulator-based ARX model, the field data-integrated GP model showed a significant 36.34\% improvement in RMSE accuracy.

Utilizing this improved HV model, the dynamic Sparse GP-based MPC policy was tested using the worldwide harmonized light vehicle test procedure (WLTP) as the reference speed profile for the leader HV. The WLTP, recognized for its realistic representation of driving conditions in car-following scenarios \cite{coppola2022eco}, makes it an ideal benchmark for evaluating the controller's performance in real-life car-following scenarios.

\begin{figure}
    \centering
    \vspace{0.15cm}
    \subfloat{{\includegraphics[trim=0.0cm 0cm 1.3cm 0cm, width=0.72\columnwidth]{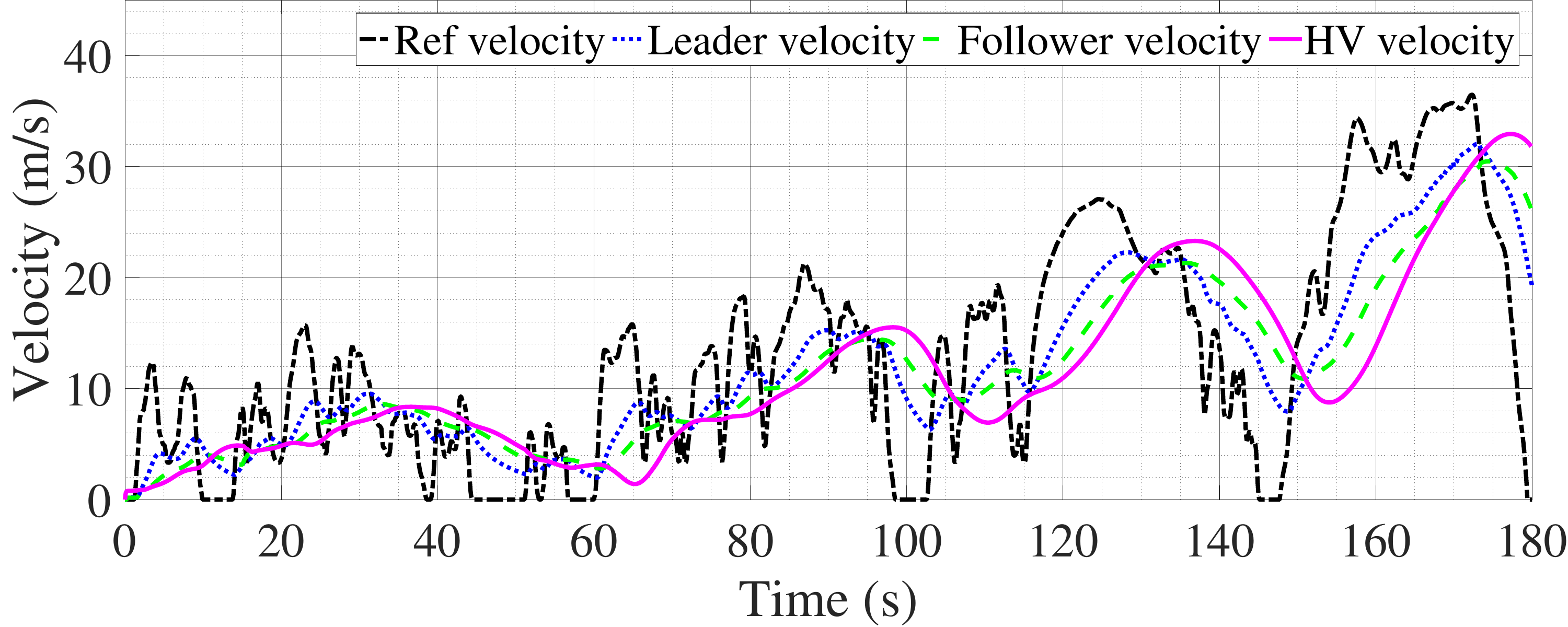} }}
    \qquad \qquad 
    \vspace{0.01cm}
    \subfloat{{\includegraphics[trim=0cm 0cm 0cm 1.3cm, width=0.76\columnwidth]{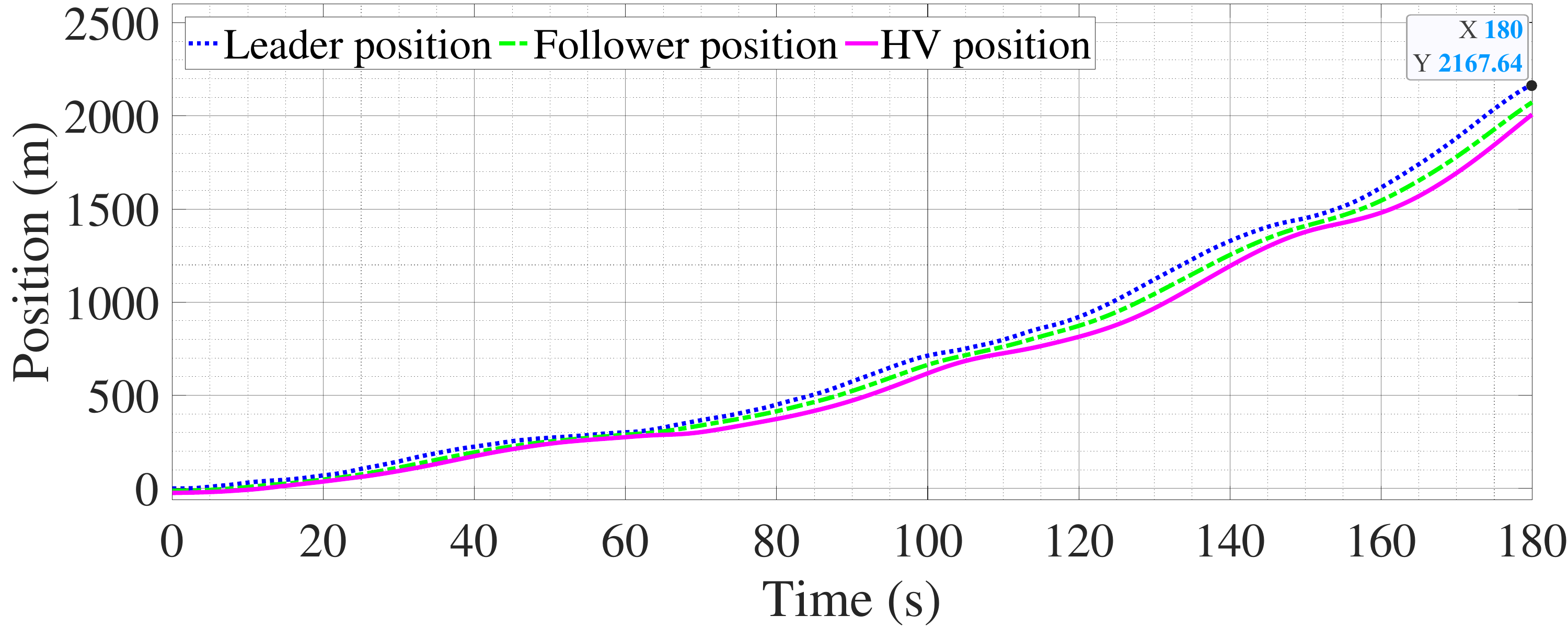} }}
    \qquad \qquad 
    \vspace{0.01cm}
    \subfloat{{\includegraphics[trim=0.0cm 0cm 1.0cm 2.0cm, width=0.74\columnwidth]{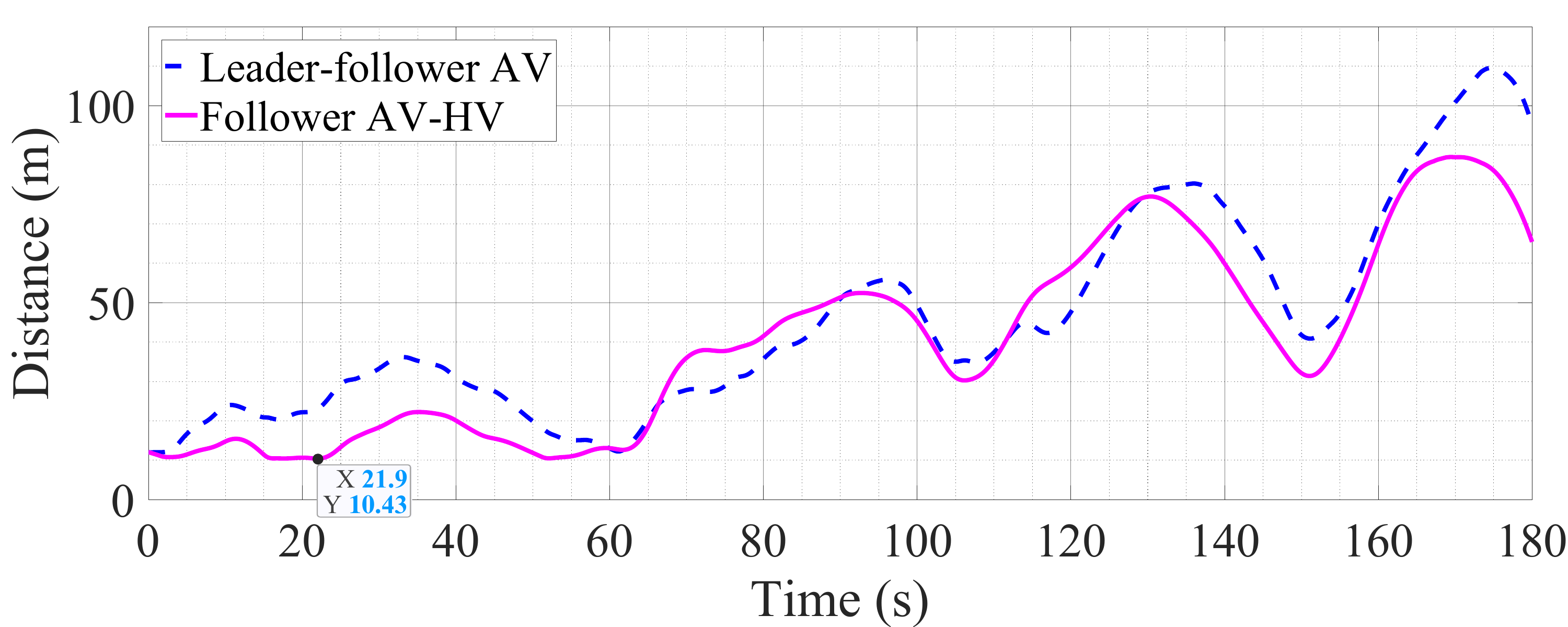} }}
    \caption{WLTP scenario trajectories under the dynamic sparse constrained GP-MPC: velocity responses, vehicle position trajectories, and inter-vehicle distances. Quantitative comparisons with the nominal-MPC baseline are reported in \cite{wang2024enhancing}.}
    \Description{WLTP scenario plots showing velocity responses, vehicle positions, and inter-vehicle distances under sparse constrained GP-MPC.}
    \label{figure:gp_simulation_wltp}
\end{figure}

Fig.~\ref{figure:gp_simulation_wltp} displays the velocity response, position trajectories, and inter-vehicle distances generated by the dynamic sparse constrained GP-MPC in the WLTP scenario. The referenced study \cite{wang2024enhancing} reports that, relative to its nominal-MPC baseline, the uncertainty-aware controller increased the minimum HV--following-AV distance by 0.4\,m while the controlled vehicle positions were 4--9\,m further ahead in the reported comparison. The figure is used here to illustrate the constrained GP-MPC trajectory behavior; the quantitative baseline comparison is attributed to the source study.
\FloatBarrier

\section{Conclusion}
\label{sec:conclusion}
This tutorial has presented a self-contained account of Gaussian process learning-based model predictive control for readers who want to understand the mathematical foundations before adapting GP-MPC to a specific implementation. After summarizing the GP residual-model notation needed for GP-MPC, the tutorial developed its central mathematical contribution: a step-by-step derivation of tractable mean and covariance propagation for GP-augmented multi-step prediction. By deriving the formulas from the laws of total expectation and total covariance, the tutorial makes explicit the roles of the GP posterior mean, posterior covariance, uncertain GP inputs, and query--output cross-covariances.

The tutorial also clarifies that GP-MPC is not a single controller formulation. Mean-only unconstrained GP predictive control uses the GP posterior mean as a learned residual correction and has the structure of deterministic predictive control. Uncertainty-aware constrained GP-MPC additionally propagates covariance and uses it to tighten probabilistic constraints. The regulation and output-tracking formulations, the reference-compatible cost, and the lifted mean-only prediction model provide a consistent notation for interpreting the examples.

This distinction makes the application sections internally consistent. The mobile-robot path-following examples are empirical demonstrations of mean-only unconstrained GP predictive control for path-error output tracking. They show how GP mean corrections can improve prediction and tracking performance, but they do not implement uncertainty-tightened constraints. The mixed-vehicle platooning example, in contrast, propagates HV-position uncertainty and uses it in a tightened safe-distance constraint together with deterministic speed and acceleration bounds; it therefore provides a concrete constrained uncertainty-aware GP-MPC implementation. The sparse variant shows how the same constrained structure can be retained while reducing the GP prediction burden.

The tutorial provides concise implementation, prototype algorithm, and complexity guidance to help readers connect the theory to practice, while emphasizing that software-level details depend on the plant model, output map, GP approximation, solver, constraints, sampling rate, and available computation. Readers seeking implementation-specific algorithms can therefore use the cited example papers as companions to the generic formulations presented here. Similarly, GP posterior uncertainty alone does not establish closed-loop stability, robust stability, recursive feasibility, or deterministic safety. Such guarantees require additional assumptions and controller-design ingredients, including calibrated uncertainty bounds, terminal or invariant-set conditions, backup strategies, or robust MPC constructions. Keeping these distinctions clear is essential for the careful development, interpretation, and evaluation of learning-enabled predictive controllers.

\section{Acknowledgments}

The authors thank Prof. Joshua A. Marshall at Queen's University for his support and mentorship of the first author. His guidance during the first author's postdoctoral tenure at Queen's University provided valuable academic direction, inspiration, and encouragement. This work has been influenced by his deep understanding of and innovative approach to robotics control. The authors also thank Prof. Yash Vardhan Pant at the University of Waterloo for his mentorship of the first author during his postdoctoral work at Waterloo, and for valuable collaboration and technical discussions on learning-based model predictive control and its applications to autonomous systems.

\appendix
\numberwithin{equation}{section}

\section{Supporting Identities for Gaussian Mean and Variance Approximation}
\label{sec:appendix_gp_approx}
This appendix summarizes the identities used to approximate GP predictions under uncertain inputs. The foundational principles of Gaussian process (GP) regression are developed in \cite{Girard2002} and \cite{girard2002gaussian}, where a GP model $f(x)$ is used to approximate a function $y = f(x)+\epsilon$. The derivation in Sec.~\ref{sec:mean_var_approx} repeatedly applies the law of total expectation and the law of total covariance. These identities are written here in a compact form so that readers can connect the main-text equations to the underlying probability rules:
\begin{align}
    m(x^*) &= \mathbb{E}_{x^*}\left[\mathbb{E}_{f(x^*)} [f(x^*) \vert x^*] \right] \, , \nonumber \\
           &= \mathbb{E}_{x^*}\left[\mu (x^*) \right] \, . \label{eqn:mean_prop_general}   \\
    \textit{var}(x^*) &= \mathbb{E}_{x^*}\left[\operatorname{var}_{f(x^*)} (f(x^*) \vert x^*) \right] + \operatorname{var}_{x^*} \left(\mathbb{E}_{f(x^*)} [f(x^*) \vert x^*] \right) , \nonumber \\
           &= \mathbb{E}_{x^*}\left[{\Sigma} (x^*) \right] + \operatorname{var}_{x^*} \left(\mu (x^*) \right) \, . \label{eqn:var_prop_general} 
\end{align}

In \eqref{eqn:mean_prop_general}, the inner conditional expectation is the GP posterior mean at a fixed query, $\mathbb{E}_{f(x^*)}[f(x^*)\vert x^*]=\mu(x^*)$. The outer expectation then accounts for the fact that the query $x^*$ itself may be uncertain. Similarly, \eqref{eqn:var_prop_general} separates the total predictive variance into the expected GP posterior variance and the variance of the GP posterior mean induced by the uncertain query. For clarity, \eqref{eqn:mean_prop_general} and \eqref{eqn:var_prop_general} are expressed explicitly with a time index:
\begin{align}
    m(x_{k+1}^*) &= \mathbb{E}_{x_k^*}\left[\mathbb{E}_{f(x^*)} [f(x^*) \vert x_{k+1}^*] \right] \, ,  \nonumber \\
     &= \mathbb{E}_{x_k^*}\left[\mu (x_{k+1}^*) \right] \, . \label{eqn:mean_prop_general_final} \\
    \textit{var}(x_{k+1}^*) &= \mathbb{E}_{x_k^*}\left[\operatorname{var}_{f(x^*)} (f(x^*) \vert x_{k+1}^*) \right] + \operatorname{var}_{x_k^*} \left(\mathbb{E}_{f(x^*)} [f(x^*) \vert x_{k+1}^*] \right) \, , \nonumber \\
    &= \mathbb{E}_{x_k^*}\left[{\Sigma} (x_{k+1}^*) \right] + \operatorname{var}_{x_k^*} \left(\mu (x_{k+1}^*) \right) \, . \label{eqn:var_prop_general_final} 
\end{align}

\subsubsection{Predictive Mean Approximation}
The approximation of the expectation under $x_k^*$ of the GP mean $\mu(x_k^*)$, as described in \eqref{eqn:mean_prop_general_final}, involves its first-order Taylor expansion around $x^*=\mu_{x_k^*}$. Following the equations in \cite[Eq. 14]{Girard2002}, this can be expressed as:
\begin{align}
    \mu(x_k^*) \approx \mu(\mu_{x_k^*}) + \frac{\partial \mu(x^*)}{\partial x^*} \bigg|_{x^*=\mu_{x_k^*}}^\top (x^* - \mu_{x_k^*}) + \mathcal{O} \left( \lvert\lvert x^* - \mu_{x_k^*} \lvert\lvert^2 \right) \, . \label{eqn:mean_taylor} 
\end{align}
Subsequently, the expectation $\mathbb{E}_{x_k^*}\left[\mu (x_{k}^*) \right]$ is approximated following the process outlined in \cite[Eq. 15]{Girard2002}:
\begin{align}
    \mathbb{E}_{x_k^*} \left[ \mu(x_k^*) \right] &\approx \mathbb{E}_{x^*_k} \left[ \mu(\mu_{x_k^*}) + \frac{\partial \mu(x^*)}{\partial x^*} \bigg|_{x^*=\mu_{x_k^*}}^\top (x^* - \mu_{x_k^*}) \right] \, , \nonumber \\
       &= \mu(\mu_{x_k^*}) \, . \label{eqn:exp_mean_appendix} 
\end{align}

\subsubsection{Predictive Variance Approximation}
Similarly, for the expectation under $x_k^*$ of the GP variance $\Sigma(x_k^*)$ in \eqref{eqn:var_prop_general_final}, we approximate $\Sigma(x_k^*)$ through its first-order Taylor expansion around $x^*=\mu_{x_k^*}$ as:
\begin{align}
    \Sigma(x_k^*) \approx {\Sigma}(\mu_{x_k^*}) + {\frac{\partial {\Sigma}(x^*)}{\partial x^*}} \bigg|_{x^*=\mu_{x_k^*}}^\top (x^* - \mu_{x_k^*}) + \mathcal{O} \left( \lvert\lvert x^* - \mu_{x_k^*} \lvert\lvert^2 \right) \, . \label{eqn:var_taylor} 
\end{align}
The expectation $\mathbb{E}_{x_k^*}\left[\Sigma (x_{k}^*) \right]$, following the derivations in \cite[Eq. 18]{Girard2002} and \cite[Appendix I]{hewing2018cautious}, is approximated as:
\begin{align}
    \mathbb{E}_{x_k^*} \left[ {\Sigma}(x_k^*) \right] & \approx \mathbb{E}_{x^*_k} \left[ {\Sigma}(\mu_{x_k^*}) + {\frac{\partial {\Sigma}(x^*)}{\partial x^*}} \bigg|_{x^*=\mu_{x_k^*}}^\top (x^* - \mu_{x_k^*}) \right] \nonumber \\
       & = {\Sigma}(\mu_{x_k^*}) \, . \label{eqn:exp_var_appendix} 
\end{align}
For the second term of \eqref{eqn:var_prop_general}, $\operatorname{var}_{x^*} \left(\mu (x^*) \right)$ is derived as follows, based on \cite[Eq. 20]{Girard2002}:
\begin{align}
    \operatorname{var}_{x_k^*} \left(\mu(x_k^*) \right) 
       &\approx \operatorname{var}_{x_k^*} \left[ \mu(\mu_{x_k^*}) + \frac{\partial \mu(x^*)}{\partial x^*} \bigg|_{x^*=\mu_{x_k^*}}^\top (x^* - \mu_{x_k^*}) \right] \nonumber \\
       &= \frac{\partial \mu(x^*)}{\partial x^*} \bigg|_{x^*=\mu_{x_k^*}} \Sigma_{x^*_k} \frac{\partial \mu(x^*)}{\partial x^*} \bigg|_{x^*=\mu_{x_k^*}} ^\top \, . \label{eqn:var_mean_appendix} 
\end{align}

\section{Reference-Compatible Quadratic Cost for Mean-Only Tracking}
\label{sec:appendix_cost_function_solve}
This appendix provides a compact closed-form solution for the unconstrained linearized mean-only tracking problem discussed in Sec.~\ref{sec:gp_mpc}. Let the lifted linearized output prediction be written as
\begin{equation}
    \delta\mathbf{y}_k = \mathcal{A}_y\delta x_{0|k}+\mathcal{B}_y\delta\mathbf{u}_k,
    \qquad
    \delta\mathbf{u}_k = \mathbf{u}_k-\mathbf{u}_{r,k},\label{eqn:appendix_lifted_output}
\end{equation}
where $\mathbf{u}_{r,k}$ is a feasible reference-input sequence. Define the nominal output-reference mismatch before applying $\delta\mathbf{u}_k$ as
\begin{equation}
    \mathbf{e}_{0,k}=\mathcal{A}_y\delta x_{0|k}-\delta\mathbf{r}_k.\label{eqn:appendix_error}
\end{equation}
For clarity, consider the quadratic cost with output tracking and reference-input deviation terms,
\begin{equation}
    J(\delta\mathbf{u}_k)
    = (\mathbf{e}_{0,k}+\mathcal{B}_y\delta\mathbf{u}_k)^{\top}\mathbf{Q}
      (\mathbf{e}_{0,k}+\mathcal{B}_y\delta\mathbf{u}_k)
      + \delta\mathbf{u}_k^{\top}\mathbf{R}\delta\mathbf{u}_k .\label{eqn:appendix_cost}
\end{equation}
An incremental-input penalty can be incorporated by augmenting the quadratic form with the corresponding lifted difference operator. For symmetric positive semidefinite $\mathbf{Q}$ and positive definite $\mathbf{R}$, setting the derivative to zero gives
\begin{equation}
    \frac{\partial J}{\partial\delta\mathbf{u}_k}
    =2(\mathcal{B}_y^{\top}\mathbf{Q}\mathcal{B}_y+\mathbf{R})\delta\mathbf{u}_k
     +2\mathcal{B}_y^{\top}\mathbf{Q}\mathbf{e}_{0,k}=0,\label{eqn:appendix_gradient}
\end{equation}
with solution
\begin{equation}
    \delta\mathbf{u}_k^{\star}
    =-(\mathcal{B}_y^{\top}\mathbf{Q}\mathcal{B}_y+\mathbf{R})^{-1}
      \mathcal{B}_y^{\top}\mathbf{Q}\mathbf{e}_{0,k},
    \qquad
    \mathbf{u}_k^{\star}=\mathbf{u}_{r,k}+\delta\mathbf{u}_k^{\star}.\label{eqn:appendix_solution}
\end{equation}
When the regulated output is an error coordinate with zero target and its compatible reference input is also zero, this formulation reduces to an absolute-input penalty. For general nonzero regulation or tracking references, retaining $\mathbf{u}_{r,k}$ avoids incorrectly driving the input toward zero and thereby introducing an avoidable steady-state offset.

\bibliographystyle{ACM-Reference-Format}
\bibliography{bibliography}

\end{document}